\pdfoutput=1
\documentclass[10pt, logo, onecolumn, copyright, nonumbering]{nvidiatechreport}

\usepackage{adjustbox} 
\usepackage{algorithm} 
\usepackage[noend]{algpseudocode} 
\usepackage{amsfonts} 
\usepackage{amsmath} 
\usepackage{amssymb} 
\usepackage{amsthm} 
\usepackage{booktabs} 
\usepackage{caption} 
\usepackage{colortbl} 
\usepackage{enumitem} 
\usepackage{float} 
\usepackage[T1]{fontenc} 
\usepackage{graphicx} 
\usepackage{hyperref} 
\usepackage[utf8]{inputenc} 
\usepackage{listings} 
\usepackage{makecell} 
\usepackage{mdframed} 
\usepackage{microtype} 
\usepackage{multicol} 
\usepackage{multirow} 
\usepackage[numbers,compress]{natbib} 
\usepackage{nicefrac} 
\usepackage{soul} 
\usepackage{stfloats} 
\usepackage{subcaption} 
\usepackage{tcolorbox} 
\usepackage{tikz} 
\usepackage{times} 
\usepackage{url} 
\usepackage{wrapfig} 
\usepackage{xspace} 
\usepackage{longtable}
\usepackage{booktabs}
\usepackage[dvipsnames]{xcolor} \usetikzlibrary{arrows.meta,decorations.pathreplacing,positioning} 

\newcommand{\armBase}{$2\times$TriMul $+$ $2\times$TriAttn\xspace} 
\newcommand{\armFull}{$4\times$TriHyena\xspace} 
\newcommand{\armHyb}{$2\times$TriMul $+$ $2\times$TriHyena\xspace} 

\newcommand{\softmax}{\operatorname{softmax}} 

\newtheorem{theorem}{Theorem} 
\newtheorem{corollary}[theorem]{Corollary} 
\newtheorem{definition}[theorem]{Definition} 
 
\newtheorem{proposition}[theorem]{Proposition} \newtheorem{remark}[theorem]{Remark}

\tcbuselibrary{listings,breakable}

\definecolor{pastelblue}{RGB}{173,216,230}
\definecolor{pastelyellow}{RGB}{255,253,208}
\definecolor{pastelpink}{RGB}{255,209,220}
\definecolor{pastelgreen}{RGB}{176,226,172}
\definecolor{pastellavender}{RGB}{230,230,250}
\definecolor{NvidiaGreen}{RGB}{118, 185, 0}
\sethlcolor{red!15}

\makeatletter
\renewcommand{\maketitle}{\bgroup\setlength{\parindent}{0pt}
        \begin{adjustwidth}{0pt}{0pt}
                \begin{flushleft}
                        {
                                {\raggedright \titlefont \@title\par}%
                                \vskip11pt
                                {\raggedright \@author\par}
                                \vskip20pt%
                        }%
                \end{flushleft}
        \end{adjustwidth}
        \egroup
        {%
                {\abscontent}
        }%
        \thispagestyle{firststyle}
}
\makeatother

\author{
  \centering
  {\small\bfseries Alireza Moradzadeh$^1$, Ben Fauber$^1$,  Youhan Lee$^1$, Nathaniel Corley$^2$, Rohith Krishna$^2$, David Baker$^2$, Michelle Gill$^1$, Emine Kucukbenli$^1$, Saee Paliwal$^1$, Olivia Viessmann$^1$}\\
  $^1$ NVIDIA\\
  $^2$ Institute for Protein Design, University of Washington, Seattle, WA, USA\\
}

\begin{document}
\title{\centering{Triangles Are All You Need: Subcubic Protein Structure Prediction with TriHyena}}

\begin{abstract}
\textbf{Abstract:}
Triangle attention (TriAttn) and triangle multiplication (TriMul) are core operations in protein structure models, but their cubic scaling limits training and inference on large biomolecules. We introduce TriHyena, a gated FFT-based triangle mixer with $O(L^2 \log L)$ compute complexity. At sequence length 2048, TriHyena trains $3.2-7.0\times $ faster than TriAttn and $1.5-1.7\times $ faster than TriMul across NVIDIA B300, B200, H200, H100, and A100 GPUs, with further gains at wider pair representations. In reduced-scale RoseTTAFold3 (RF3) ablations, replacing TriAttn with TriHyena preserves baseline protein and protein-protein interface lDDT. Replacing both TriAttn and TriMul with TriHyena, yielding an all-TriHyena model, retains baseline lDDT for monomers and is close in lDDT (-0.04) for protein interfaces. In protein generation with Genie2, replacing TriMul with TriHyena yields comparable designability and diversity. TriHyena is thus an efficient substitute for TriAttn and TriMul, and a basis for architectures that further reduce computational cost. Code: \url{https://github.com/NVIDIA-BioNeMo/nvSubquadratic}. 
\end{abstract}

\maketitle

\section{Introduction}
\label{sec:intro}

Protein structure prediction models update pair representations using triangle attention (TriAttn) \citep{bergen2021edgetransformers,jumper2021alphafold} and triangle multiplication (TriMul) \citep{jumper2021alphafold}. Both mix information over three-residue triangles, capturing geometric relationships between residue pairs. Together, they form the pair stack of AlphaFold2/3 \citep{jumper2021alphafold,abramson2024alphafold3}, Boltz1/2 \citep{wohlwend2024boltz1,passaro2025boltz2}, RosettaFold-3 (RF3) \citep{corley2025rf3}, and subsequent models \citep{baek2023rf2,chai2024chai1,bytedance2025protenix,zhang2026protenixv2,zhang2026protenixv1}, with ablations demonstrating the importance of triangle mixing for accuracy.

Triangle mixing comes at substantial computational cost: each of $L^2$ residue pairs (where $L$ is the number of tokens) aggregates information across the sequence, giving both operators $O(L^3)$ time and $O(L^2)$ memory complexity and limiting training and high-throughput inference on large complexes. Existing approaches reduce these costs but retain fundamental trade-offs. Optimized kernels \citep{song2023deepspeed4scienceinitiativeenablinglargescale,anthropic_flashpairformer} and TriMul-only architectures \citep{ouyang-zhang2026triangle, geffner2025proteina} reduce computational constants but remain cubic, while subcubic approaches approximate or remove explicit triangle mixing, potentially sacrificing accuracy or triangular structure \citep{candido2026language,qiang2025protenixminiplus}. Preserving explicit triangle mixing while breaking the cubic time barrier therefore remains an open challenge.

We introduce Triangle Hyena (TriHyena), a triangle mixing operator that preserves this structure while reducing time complexity from $O(L^3)$ to $O(L^2 \log L)$, a $\Theta(L/\log L)$ asymptotic speedup, with unchanged $O(L^2)$ memory complexity. TriHyena achieves this scaling by mixing gated pair features through learned FFT convolutions.

\section{Related Work}
\label{sec:related}

\begin{figure}[t]
\centering
\begin{tikzpicture}[
  y=0.85cm,
  >={Stealth[length=7pt, width=6.5pt]},
  text=black,
  qnode/.style={circle, draw=black!60, fill=black!5, thick,
                minimum size=0.72cm, font=\small, inner sep=0},
  knode/.style={circle, draw=NvidiaGreen, fill=NvidiaGreen!12, thick,
                minimum size=0.72cm, font=\small, inner sep=0},
  kedge/.style={NvidiaGreen, thick},
  koff/.style={->, NvidiaGreen, thick, dashed},
  bedge/.style={black!60, thick},
  elab/.style={font=\scriptsize, text=black, sloped, inner sep=1.5pt},
  note/.style={font=\scriptsize, text=black!60, align=center},
  title/.style={font=\small},
]

\def\tri#1#2{
  \node[knode] (#2k) at (#1+1.4, 4.052) {$k$};
  \node[qnode] (#2i) at (#1+0.1, 1.80) {$i$};
  \node[qnode] (#2j) at (#1+2.7, 1.80) {$j$};
}

\node[font=\small\bfseries, anchor=west] at (-0.25, 4.75) {(A)};
\node[title] at (1.4, 4.75) {TriAttn};
\tri{0}{a}
\draw[kedge] (ai) -- (ak) node[elab, midway, above] {key + value};
\draw[kedge] (aj) -- (ak) node[elab, midway, above] {attn bias};
\draw[bedge] (ai) -- (aj) node[elab, midway, below] {query + gate};
\node[note] at (1.4, 1.05) {softmax weights over $k$};

\node[font=\small\bfseries, anchor=west] at (3.95, 4.75) {(B)};
\node[title] at (5.6, 4.75) {TriMul};
\tri{4.2}{m}
\draw[kedge] (mi) -- (mk) node[elab, midway, above] {$\mathbf{p}_{i,k}$};
\draw[kedge] (mj) -- (mk) node[elab, midway, above] {$\mathbf{q}_{j,k}$};
\draw[bedge] (mi) -- (mj) node[elab, midway, below] {output gate};
\node[note] at (5.6, 1.05) {uniform weights};

\node[font=\small\bfseries, anchor=west] at (8.15, 4.75) {(C)};
\node[title] at (9.8, 4.75) {TriHyena};
\tri{8.4}{h}
\draw[kedge] (hi) -- (hk)
  node[elab, midway, above] {$\mathbf{U}_{i,k}\odot\mathbf{U}_{k,i}$};
\draw[koff] (hk) -- (hj) node[elab, midway, above] {offset $j-k$};
\draw[bedge] (hi) -- (hj) node[elab, midway, below] {output gate};
\node[note] at (9.8, 1.05) {learned $\mathbf{h}_{j-k}$ weights};

\end{tikzpicture}
\caption{%
  Triangle mixing geometries depict three mixers updating the target pair $(i,j)$ by aggregating over $k$: grey marks target-pair inputs and gates, green edges mark pair features, and the dashed edge marks only a sequence offset.
  (A) TriAttn reads keys and values from $(i,k)$ and a bias from $(j,k)$, using input-dependent softmax weights over $k$.
  (B) TriMul multiplies projections of $(i,k)$ and $(j,k)$ elementwise and sums them uniformly over $k$.
  (C) TriHyena filters reciprocal $(i,k)$ and $(k,i)$ features with learned depthwise weights $\mathbf{h}_{j-k}$ via FFT convolution, without reading $(j,k)$ features.
}
\label{fig:geometry}
\end{figure}
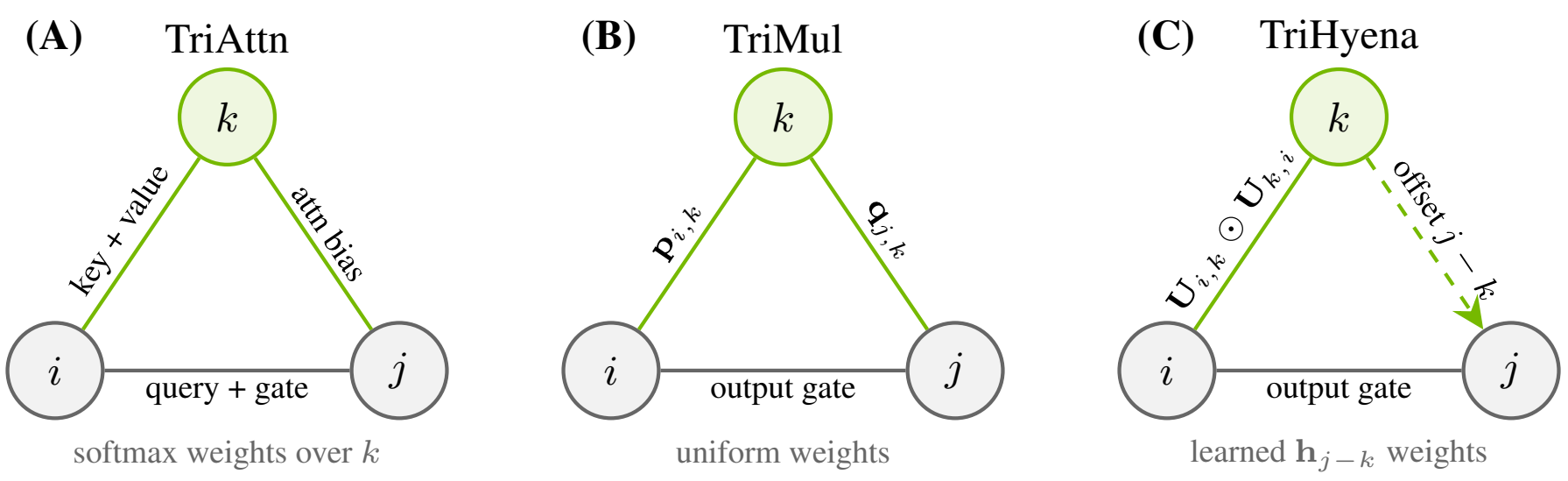

\paragraph{Triangle mixing in biomolecular structure prediction.}
AlphaFold-style models update pair representations using two operators that share the same triangle geometry (Figure~\ref{fig:geometry}). TriAttn \citep{bergen2021edgetransformers,jumper2021alphafold} updates each pair by attending to pairs that share one endpoint, with attention biased by the third edge completing the triangle. TriMul \citep{jumper2021alphafold} instead combines projected features from the other two edges to update the target pair. Both require $O(L^3)$ time and $O(L^2)$ memory, although TriMul has a smaller computational constant because its contraction reduces to dense matrix multiplication. Together, TriAttn and TriMul form the pair stack of most subsequent structure prediction models \citep{abramson2024alphafold3,wohlwend2024boltz1,passaro2025boltz2,corley2025rf3,chai2024chai1,bytedance2025protenix,zhang2026protenixv2,zhang2026protenixv1,lin2023esmfold}.

\paragraph{Accelerating, approximating, or removing the triangle stack.}
Three approaches mitigate the cubic cost of triangle mixing. First, optimized implementations preserve the exact operators: fused and tiled kernels such as \texttt{cuEquivariance}\footnote{\url{https://github.com/nvidia/cuequivariance} (accessed 14Oct2026)}, DS4Sci\_EvoformerAttention \citep{song2023deepspeed4scienceinitiativeenablinglargescale}, and FlashPairformer \citep{anthropic_flashpairformer} reduce wall time and memory to $O(L^2)$ while retaining $O(L^3)$ computation. Second, cheaper substitutes reduce or approximate the triangle operators. Pairmixer \citep{ouyang-zhang2026triangle} removes TriAttn but retains cubic TriMul, whereas Protenix-Mini+ \citep{qiang2025protenixminiplus} achieves quadratic cost by approximating TriAttn with kernelized linear attention and applying chunked TriMul in a subset of blocks, with a reported loss in interface accuracy. Third, some methods remove triangle mixing entirely in favor of non-equivariant transformers or protein language-model representations \citep{candido2026language, wang2025simplefold}, sacrificing the triangle inductive bias. La-Proteina omits triangular update layers in its main models for scalability, and reports that adding TriMul improves co-designability \citep{geffner2025proteina}. In contrast, TriHyena introduces a distinct triangle operator that preserves three-index geometry at subcubic cost.

\paragraph{Subquadratic mixers via long convolutions.}
A broad family of operators performs global sequence mixing without materializing dense pair representations. State-space models such as S4 \citep{gu2022s4} and Mamba \citep{gu2024mamba} use structured recurrences, with Mamba adding input-dependent gating. Hyena \citep{poli2023hyena} instead uses implicitly parameterized long convolutions computed efficiently with FFTs, coupled with multiplicative gates. HyenaDNA \citep{nguyen2023hyenadna}, defined for $1\mathrm{D}$ sequences, extends this approach to genomic sequences of up to $10^6$ tokens.

This long-convolution framework also extends beyond $1\mathrm{D}$ sequences. HyenaND \citep{wessels2026} generalizes Hyena to the native geometry of multidimensional data, such as images and volumes, using implicitly parameterized multidimensional kernels. It also provides the fused FFT-convolution CUDA library \texttt{nvSubquadratic}\footnote{\url{https://github.com/NVIDIA-BioNeMo/nvSubquadratic} (accessed 14Oct2026)}, translating the favorable $O(L\log L)$ scaling of FFT convolution into wall-clock speedups. TriHyena builds on this multidimensional long-convolution framework: it updates the 2D pair representation, replacing the cubic summation over the shared triangle index with a $1\mathrm{D}$ FFT convolution over the ordered sequence, while using the same fused-FFT kernel infrastructure.
Unlike a standard 2D HyenaND applied to the $L\times L$ pair grid, which would convolve over both pair indices, TriHyena is the $n{=}3$ instance of an $n$-index Hyena (Appendix~\ref{app:attn-n}): it convolves only along the shared triangle index $k$ and gates all three edges of the triangle (Section~\ref{sec:method}).

\section{Background}
\label{sec:background}

In biomolecular structure prediction and design architectures, each residue pair $(i,j)$ is represented by a learned feature vector, yielding a pair tensor $\mathbf{U}\in\mathbb{R}^{B\times L\times L\times d_{\mathrm{pair}}}$, where $B$ is the batch size, $L$ the sequence length, and $d_{\mathrm{pair}}$ the pair channel dimension.  All three triangle mixers update pair $(i, j)$ by aggregating over the shared index $k$, gating the result, and projecting back to the pair width.

\subsection{Triangle Attention}
\label{sec:triattn}
For TriAttn (starting node version) \citep{bergen2021edgetransformers, jumper2021alphafold}, each pair $(i,j)$ attends over pairs $(i,k)$ along row $i$ (Figure \ref{fig:geometry}) through the core operation: 

\begin{equation} 
\mathrm{TriAttn}(\mathbf{U})_{i,j} = \sum_{k=1}^{L} A_{i,j,k}\mathbf{V}^{\mathrm{att}}_{i,k}, \qquad A_{i,j,k} = \softmax_{k} \left( \frac{\mathbf{Q}_{i,j}^{\mathsf{T}}\mathbf{K}_{i,k}} {\sqrt{d_k}} + b_{j,k} \right),
\label{eq:triattn} 
\end{equation}

where $\mathbf{Q}$, $\mathbf{K}$, and $\mathbf{V}^{\mathrm{att}}$ are learned projections of the layer-normalized pair representation to $\mathbb{R}^{d_k}$ per head, and $b_{j,k}$ is the attention bias, a per-head linear projection of $\mathbf{U}_{j,k}$ shared across rows $i$. Eq.~\eqref{eq:triattn} is written for one head and omits the sigmoid output gate for brevity. TriAttn uses $n_{\mathrm{head}}$ heads of width $d_k{=}d_{\mathrm{hidden}}$, whose outputs are concatenated, gated, and projected back to the pair width by $W_{\mathrm{out}}\in\mathbb{R}^{n_{\mathrm{head}}d_{\mathrm{hidden}}\times d_{\mathrm{pair}}}$. Each of the $L^2$ output pairs aggregates over $L$ values of $k$, giving $O(L^3)$ time complexity. This cubic scaling makes TriAttn increasingly costly for long sequences and large complexes.

\subsection{Triangle Multiplication}
\label{sec:trimul}
For TriMul (outgoing variant) \citep{jumper2021alphafold}, each pair $(i,j)$ aggregates multiplicative interactions between pairs $(i,k)$ and $(j,k)$ over the shared index $k$: 

\begin{equation} 
\mathrm{TriMul}(\mathbf{U})_{i,j} = g_{i,j} \odot \left(\sum_{k=1}^{L} p_{i,k} \odot q_{j,k}\right)W_{\mathrm{out}}, \quad
p_{i,k} = \sigma(\mathbf{U}_{i,k}W_{gp}) \odot (\mathbf{U}_{i,k}W_{p}), \;\;
q_{j,k} = \sigma(\mathbf{U}_{j,k}W_{gq}) \odot (\mathbf{U}_{j,k}W_{q}),
\label{eq:trimul} 
\end{equation} 

where $W_{p},W_{q},W_{gp},W_{gq}\in\mathbb{R}^{d_{\mathrm{pair}}\times c}$ project to the internal channel width $c$, $W_{\mathrm{out}}\in\mathbb{R}^{c\times d_{\mathrm{pair}}}$ is the learned output projection, $g_{i,j}=\sigma(\mathbf{U}_{i,j}W_{g})$ with $W_{g}\in\mathbb{R}^{d_{\mathrm{pair}}\times d_{\mathrm{pair}}}$ is the output gate, $\sigma$ denotes the sigmoid function, and $\odot$ denotes elementwise multiplication across channels. We take $c{=}d_{\mathrm{pair}}$ throughout, as required by cuEquivariance. The implementation additionally applies layer normalization to $\mathbf{U}$ on input and to the aggregated sum before $W_{\mathrm{out}}$; both are omitted from Eq.~\eqref{eq:trimul} for brevity. TriMul shares the $O(L^3)$ time complexity of TriAttn but avoids attention and softmax over the shared index $k$, giving it a smaller computational constant in practice.

\begin{table}[t] 
\centering \footnotesize \setlength{\tabcolsep}{4pt} 
\begin{tabular}{llcc} 
\toprule $n$ & Operator & Time & Memory \\ 
\midrule 
2 & Self-attn & $O(L^2)$ & $O(L)$ \\ 
2 & Hyena & $O(L\log L)$ & $O(L)$ \\ 
3 & TriAttn & $O(L^3)$ & $O(L^2)$ \\ 
3 & TriMul & $O(L^3)$ & $O(L^2)$ \\ 
3 & TriHyena & $O(L^2\log L)$ & $O(L^2)$ \\ 
\bottomrule 
\end{tabular} 
\caption{Time and memory complexity of self-attention \citep{vaswani2017attention}, Hyena/HyenaDNA \citep{nguyen2023hyenadna}, TriAttn and TriMul \citep{jumper2021alphafold}, and TriHyena (ours). Memory assumes memory-efficient kernels; channel widths omitted.} 
\label{tab:complexity} 
\end{table}

\subsection{Hyena Operators}
\label{sec:hyena2}
The Hyena operator \citep{poli2023hyena} combines learned convolutions with data-dependent multiplicative gates. We consider the two-gate, one-long-convolution specialization and generalize the number of indices interacting at each step. Let $n \geq 2$ denote this number. The input and output retain $n-1$ spatial indices, $\mathbf{U}, \mathbf{Y} \in \mathbb{R}^{B \times L_1 \times \cdots \times L_{n-1} \times d_{\mathrm{pair}}}$, while convolution along an active axis $\alpha \in [n-1]$ introduces one additional source index. Here $d_{\mathrm{pair}}$ is the input and output channel width, and $D$ is the internal Hyena channel width, a hyperparameter that the operator does not constrain relative to $d_{\mathrm{pair}}$ (our implementation sets $D{=}d_\mathrm{pair}{=}n_{\mathrm{head}}d_{\mathrm{hidden}}$, unlike TriAttn the operator has no further head structure: the filter is depthwise, so only $D$ enters). This generalized Hyena operator can be expressed as the following five-stage computation:

\begin{enumerate}[leftmargin=*, itemsep=1pt, topsep=2pt] 
    \item \textbf{Projection and local mixing:} The normalized input is projected to $3D$ channels, potentially from different shape-compatible index views, then locally mixed along axis $\alpha$ by a short depthwise convolution.
    \item \textbf{Channel split:} The result is split into two gate streams and one value stream, $\mathbf{X}^{(1)},\mathbf{X}^{(2)},\mathbf{V} \in \mathbb{R}^{B \times L_1 \times \cdots \times L_{n-1} \times D}$. 
    \item \textbf{First gate:} $\mathbf{Z} = \mathbf{X}^{(1)} \odot \mathbf{V}$.
\item \textbf{Global mixing:} A learned long filter $\mathbf{h}$ (one filter per channel) mixes the gated features along axis $\alpha$ via FFT convolution: $\mathbf{Z}' = \mathbf{h} \ast_{\alpha} \mathbf{Z}$.
    \item \textbf{Second gate and output:} $\mathbf{Y} = \left(\mathbf{X}^{(2)} \odot \mathbf{Z}'\right) W_{\mathrm{out}}$, where $W_{\mathrm{out}}\in\mathbb{R}^{D\times d_{\mathrm{pair}}}$ is a learned affine output projection, applied by right-multiplication on the channel axis independently at each retained index tuple.
\end{enumerate} 

Because convolution mixes information across positions along the active axis, causality determines whether future positions can influence the output. Causal mixing restricts a convolution to the current and preceding positions, preventing future-information leakage in autoregressive tasks; non-causal mixing may use both directions when the full input is available.

For $n{=}2$, the short convolution in stage~1 provides local mixing in $O(L)$ time, while the FFT-based convolution in stage~4 provides global mixing in $O(L\log L)$ time, compared with $O(L^2)$ for dense self-attention. TriHyena generalizes this sequence construction to $n{=}3$ interactions over pair representations with two retained index axes. The general $n$-index case is detailed in Appendix~\ref{app:nindex}.

\section{TriHyena}
\label{sec:method}

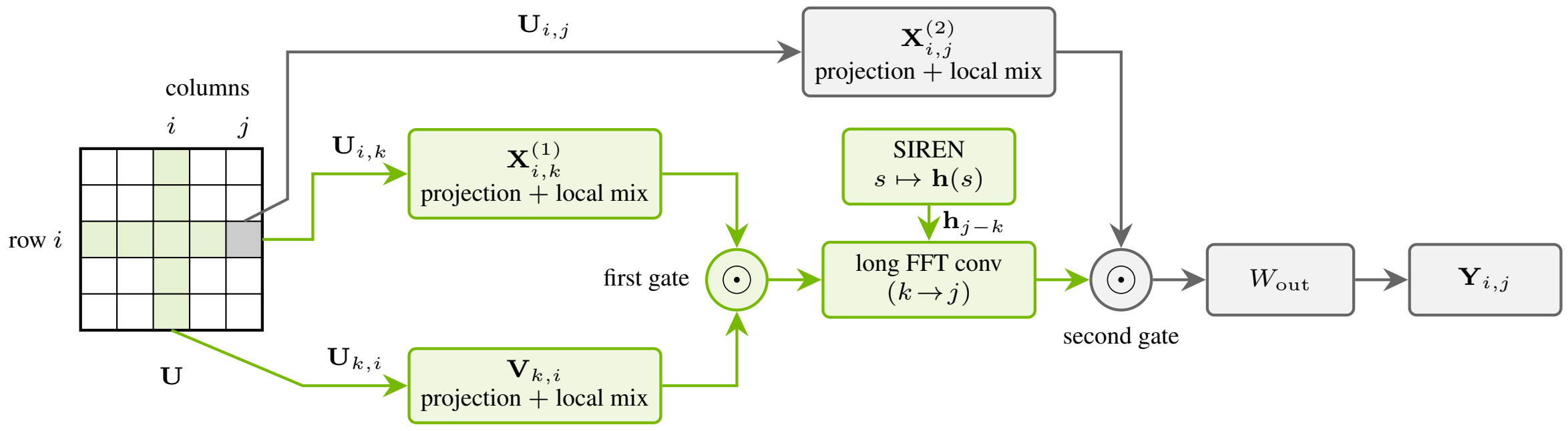
\begin{figure}[t]
\centering
\begin{tikzpicture}[
  >={Stealth[length=7pt, width=6.5pt]},
  text=black,
  stage/.style={draw=NvidiaGreen, fill=NvidiaGreen!12, thick,
    rounded corners=2pt, minimum width=1.7cm,
    minimum height=0.70cm, align=center, font=\scriptsize},
  gate/.style={circle, draw=NvidiaGreen, fill=NvidiaGreen!12,
    thick, minimum size=0.60cm, inner sep=0pt, font=\large},
  arr/.style={->, draw=NvidiaGreen, thick}
]

\begin{scope}[shift={(0,2.0)}]
  \fill[NvidiaGreen!18] (0,0.72) rectangle (1.8,1.08);
  \fill[NvidiaGreen!18] (0.72,0) rectangle (1.08,1.8);
  \fill[black!20] (1.44,0.72) rectangle (1.8,1.08);
  \draw[step=0.36cm, thin, black] (0,0) grid (1.8,1.8);
  \draw[thick, black] (0,0) rectangle (1.8,1.8);
  \node[font=\scriptsize, anchor=east] at (-0.05,0.90) {row $i$};
  \node[font=\scriptsize, anchor=south] at (1.26,2.20) {columns};
  \node[font=\scriptsize, anchor=base] at (0.90,1.95) {$i$};
  \node[font=\scriptsize, anchor=base] at (1.62,1.95) {$j$};
\end{scope}
\node[font=\scriptsize] at (0.9,1.55) {$\mathbf{U}$};

\node[stage] (xone) at (4.5,3.55)
  {$\mathbf{X}^{(1)}_{i,k}$\\{\scriptsize projection $+$ local mix}};
\node[stage] (value) at (4.5,1.45)
  {$\mathbf{V}_{k,i}$\\{\scriptsize projection $+$ local mix}};
\node[gate] (pre) at (6.5,2.5) {$\odot$};
\node[stage, minimum width=2.10cm, right=0.525cm of pre] (fft)
  {long FFT conv\\$(k\!\to\!j)$};
\node[stage, above=0.35cm of fft] (siren)
  {SIREN\\$s\mapsto\mathbf{h}(s)$};
\node[stage, draw=black!60, fill=black!5, above=0.30cm of siren] (xtwo)
  {$\mathbf{X}^{(2)}_{i,j}$\\{\scriptsize projection $+$ local mix}};
\node[gate, draw=black!60, fill=black!5, right=0.525cm of fft]
  (post) {$\odot$};
\node[stage, draw=black!60, fill=black!5, minimum width=1.45cm,
  right=0.525cm of post] (wout) {$W_{\mathrm{out}}$};
\node[stage, draw=black!60, fill=black!5, minimum width=1.50cm,
  right=0.525cm of wout] (yout) {$\mathbf{Y}_{i,j}$};
\node[left=0.03cm of pre, font=\scriptsize] {first gate};
\node[below=0.03cm of post, font=\scriptsize] {second gate};

\draw[arr] (1.8,2.9) -- (2.3,2.9) -- (2.3,3.55)
  -- node[above, font=\scriptsize] {$\mathbf{U}_{i,k}$} (xone.west);
\draw[arr, draw=black!60] (1.62,3.08) -- (2.05,3.30)
  |- node[above, near end, font=\scriptsize] {$\mathbf{U}_{i,j}$} (xtwo.west);
\draw[arr] (0.9,2.0) -- (2.2,1.45)
  -- node[above, font=\scriptsize] {$\mathbf{U}_{k,i}$} (value.west);
\draw[arr] (xone.east) -| (pre.north);
\draw[arr] (value.east) -| (pre.south);
\draw[arr] (pre.east) -- (fft.west);
\draw[arr] (siren.south) -- node[right, font=\scriptsize] {$\mathbf{h}_{j-k}$} (fft.north);
\draw[arr] (fft.east) -- (post.west);
\draw[arr, draw=black!60] (xtwo.east) -| (post.north);
\draw[arr, draw=black!60] (post.east) -- (wout.west);
\draw[arr, draw=black!60] (wout.east) -- (yout.west);
\end{tikzpicture}

\caption{TriHyena starting-node interaction for one output pair. Green represents pair-feature processing and filtering; grey represents target-pair gating and input/output. Channels are suppressed. The anchor $i$ selects a row and reciprocal column, the target $j$ identifies the output, and $k$ is swept through both. The row and column streams are multiplied, mixed from the swept positions to the target by long depthwise FFT convolution using a SIREN-generated filter, gated by the target stream, and output-projected.}
\label{fig:skeleton}
\end{figure}

\subsection{TriHyena Operator}
\label{sec:operator}
For each row index $i$, TriHyena derives gate streams $\mathbf{X}^{(1)}$ and $\mathbf{X}^{(2)}$ from pairs $\mathbf{U}_{i,k}$ and a value stream $\mathbf{V}$ from the reciprocal pairs $\mathbf{U}_{k,i}$. After local mixing, the first gate and value streams are combined elementwise, and a learned filter $\mathbf{h}$ with taps $\mathbf{h}_{j-k}\in\mathbb{R}^{D}$ mixes the resulting features along the sequence axis through FFT convolution (Figure \ref{fig:geometry}): 

\begin{equation}
\mathrm{TriHyena}(\mathbf{U})_{i,j} = \left( \mathbf{X}^{(2)}_{i,j} \odot \sum_{k=1}^{L} \mathbf{h}_{j-k} \odot \bigl( \mathbf{X}^{(1)}_{i,k} \odot \mathbf{V}_{k,i} \bigr) \right) W_{\mathrm{out}}.
\label{eq:trihyena-output} 
\end{equation} 

Here, subscripts index entries of the corresponding tensors of Section~\ref{sec:hyena2} (batch index suppressed): $\mathbf{X}^{(1)}_{i,k},\mathbf{X}^{(2)}_{i,j},\mathbf{V}_{k,i}$ and the filter taps $\mathbf{h}_{j-k}$ lie in $\mathbb{R}^{D}$, with $D{=}n_{\mathrm{head}}\,d_{\mathrm{hidden}}$ the internal channel width, and $\odot$ acts channelwise (the filter is depthwise). $W_{\mathrm{out}}\in\mathbb{R}^{D\times d_{\mathrm{pair}}}$ denotes the learned output projection (with bias), applied by right-multiplication independently at each pair $(i,j)$. The layer does not constrain $D$ relative to $d_{\mathrm{pair}}$; the widths used are stated with each experiment in Section~\ref{sec:experiments}. 

Figure~\ref{fig:skeleton} summarizes the implementation flow. Equation~\eqref{eq:trihyena-output} also highlights that TriHyena is not simply an FFT implementation of the TriAttn operator in Equation~\eqref{eq:triattn}. In the starting-node orientation, $\mathbf{X}^{(1)}_{i,k}$ is derived from $\mathbf{U}_{i,k}$, while $\mathbf{V}_{k,i}$ is derived from the reciprocal pair $\mathbf{U}_{k,i}$. Their gated product is transported from $k$ to $j$ according to the sequence-relative offset $j-k$ and modulated by the second gate $\mathbf{X}^{(2)}_{i,j}$. The ending-node orientation applies the analogous operation around $j$. TriHyena therefore preserves three-index interactions while replacing attention over the shared index with gated, sequence-relative convolution.

\subsection{Computational and Mathematical Distinction} 
\label{sec:complexity} 
TriHyena requires $O(L^2\log L)$ time and $O(L^2)$ memory (Table~\ref{tab:complexity}). Following the convention of Section~\ref{sec:hyena2}, stages~1, 3, and~5 of TriHyena operate pointwise over the $L^2$ pair cells, while stage~4 applies a length-$L$ FFT convolution to each of the $L$ rows. The resulting cost is $O(L \cdot L\log L)=O(L^2\log L)$. Memory complexity is $O(L^2)$, dominated by the $L \times L \times d_{\mathrm{pair}}$ pair representation. TriHyena achieves a $\Theta(L/\log L)$ asymptotic speedup over the $O(L^3)$ TriAttn and TriMul. Appendix~\ref{app:complexity} provides the full complexity analysis and establishes the corresponding tight bounds.

TriAttn, TriMul, and TriHyena define mathematically distinct operator families. TriAttn mixes over the shared index $k$ with input-dependent softmax weights; TriMul multiplicatively combines $(i,k)$ and $(j,k)$ and sums uniformly over $k$; and TriHyena gates $(i,k)$ and $(k,i)$, convolves from $k$ to $j$ with a row-shared filter depending only on $j-k$, and gates the output at $(i,j)$. TriHyena ($n{=}3$) is a mathematically distinct operator (Appendix~\ref{app:proofs}) that shares the input and output shape of TriAttn and TriMul and can substitute for either, but it must be trained from scratch. Models trained with different mixing methods have non-transferable weights and must use the same method at inference. 

\begin{figure}[t]
\centering
\includegraphics[width=0.75\linewidth]{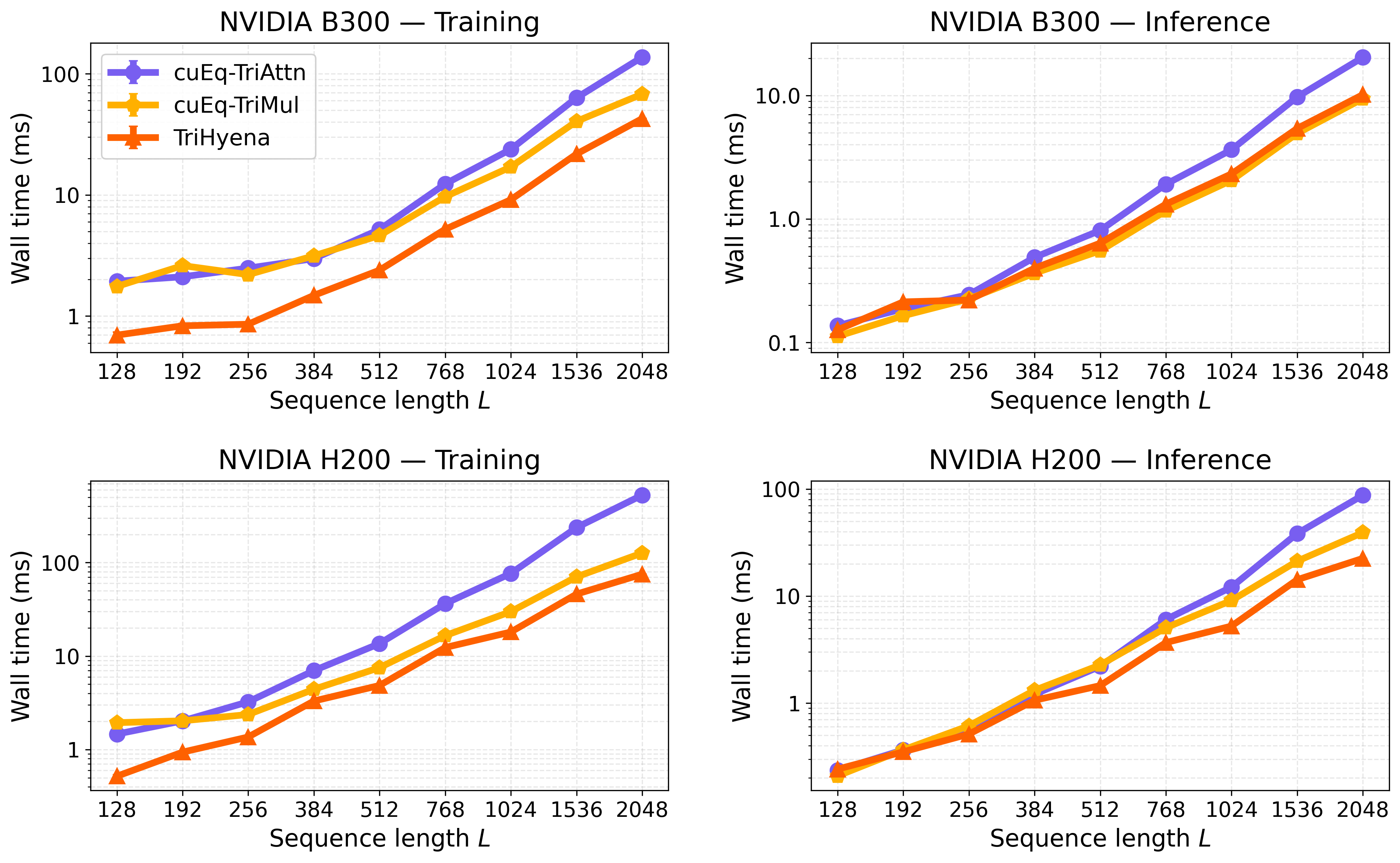}
\caption{Training (forward + backward) and inference (forward only) wall time vs. sequence length for cuEq-TriAttn (blue), cuEq-TriMul (gold), and TriHyena (orange) on NVIDIA B300 and H200 GPUs (bf16, $B{=}1$, $d_{\mathrm{pair}}{=}256$, $n_{\mathrm{head}}{=}4$). Points show the mean of five replicates; error bars indicate $\pm 1$ standard deviation.}
\label{fig:b300-h200-facet}
\end{figure}

\subsection{Implementation and Optimization for the Protein Domain} 
\label{sec:implementation} 
Algorithm~\ref{alg:triangle_hyena} provides forward-pass pseudocode (Appendix~\ref{app:implementation}). The global filter $\mathbf{h}$ is parameterized by a sinusoidal implicit network (SIREN) \citep{poli2023hyena,sitzmann2020siren}, mapping the normalized offset $s=(j-k)/(L_{\mathrm{cache}}-1)$ to $D$ channels with $2L-1$ taps per channel. Global mixing uses $1\mathrm{D}$ FFT convolution along the sequence axis, with padded positions zeroed in both gate streams. The short convolution is causal, whereas the global filter spans offsets $j-k\in[-(L-1),L-1]$ and is two-sided.

\paragraph{Optimizing for protein sequence lengths.}
Hyena was developed for long-context applications \citep{poli2023hyena}, including genome-scale sequences \citep{nguyen2023hyenadna}, where $O(L\log L)$ FFT convolution provides a substantial asymptotic advantage. However, protein structure prediction typically uses shorter sequences ($L\leq 2048$), where launch overhead, HBM traffic, and hardware utilization strongly affect wall-clock performance. We therefore implement complementary FFT- and Tensor Core-based convolution paths. Fused cuFFTDx\footnote{\url{https://docs.nvidia.com/cuda/cufftdx/} (accessed 14Oct2026)} kernels combine the forward FFT, filter multiplication, inverse FFT, and output extraction, reducing launches and intermediate HBM traffic, with block and elements-per-thread configurations tuned by sequence length and GPU architecture.

On NVIDIA B200 and B300 GPUs, we additionally implement explicit convolution using Tensor Core matrix multiply-accumulate (MMA) operations; other architectures and unsupported lengths use the FFT path. Although explicit convolution requires $O(L^2)$ arithmetic per sequence and channel rather than $O(L\log L)$, it can be faster for $L\leq 2048$ by exploiting Tensor Core throughput and avoiding FFT overhead. The FFT path retains TriHyena's subcubic complexity and provides the portable implementation across architectures.

Finally, the forward pass includes projections, short causal convolutions, and elementwise gates (Figure~\ref{fig:skeleton}; Algorithm~\ref{alg:triangle_hyena}). We combine compatible operations and use compiled PyTorch to fuse adjacent pointwise and gating operations, reducing launch, dispatch, and memory-traffic overhead. Together, fused FFT convolution, Tensor Core explicit convolution, and operator fusion optimize TriHyena for protein-relevant sequence lengths and hardware.

\section{Experiments}
\label{sec:experiments}

\begin{figure}
\centering
\includegraphics[width=1.0\linewidth]{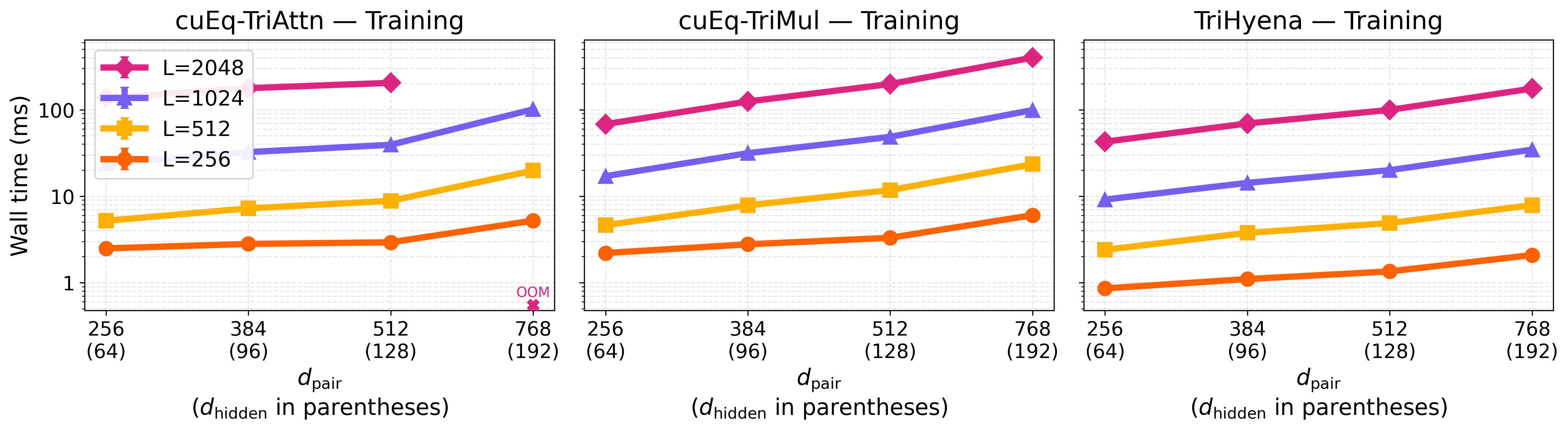}
\caption{Training (forward + backward) wall time vs. pair dimension ($d_{\mathrm{pair}}$) for cuEq-TriAttn, cuEq-TriMul, and TriHyena on an NVIDIA B300 GPU (bf16, $B{=}1$, $n_{\mathrm{head}}{=}4$). Colors denote sequence length ($L = 256, 512, 1024, 2048$); $\boldsymbol{\times}$ marks OOM. Points show the mean of five replicates; error bars indicate $\pm 1$ standard deviation.}
\label{fig:b300-training-runtime-channel-benchmark}
\end{figure}

\subsection{Benchmarking}
\label{sec:benchmarking-description}
TriAttn and TriMul are core triangle operations in structure prediction and design models. For all benchmarks, we compare TriHyena against optimized \texttt{cuEquivariance} implementations of TriAttn (cuEq-TriAttn) and TriMul (cuEq-TriMul), which substantially reduce runtime relative to their native PyTorch counterparts. All results use bfloat16 (bf16) precision for inputs and outputs. Benchmarks use $d_\mathrm{pair}{=}256$ by default (varied below) and $n_\mathrm{head}{=}4$, with the TriHyena channel width $D{=}d_\mathrm{pair}$ throughout. Results are averaged over five replicates per sample unless otherwise noted.

\subsection{TriHyena Accelerates Training}
\label{sec:trimixing-results}
We benchmark TriHyena against cuEq-TriAttn and cuEq-TriMul across seven NVIDIA GPUs spanning Blackwell, Hopper, and Ampere at sequence lengths up to $L{=}2048$ (Figures~\ref{fig:b300-h200-facet} and \ref{fig:nvidia-gpus-benchmark}). During training, TriHyena is faster than cuEq-TriAttn in all matched configurations, with a 2.8$\times$ average speedup, and faster than cuEq-TriMul in 97\% of matched configurations, with a 2.0$\times$ average speedup. On B300, TriHyena achieves 2.6$\times$ and 2.1$\times$ average speedups over cuEq-TriAttn and cuEq-TriMul, respectively; on H200, the corresponding average speedups are 3.5$\times$ and 1.9$\times$, reaching 7.0$\times$ over cuEq-TriAttn at $L{=}2048$. Averages reported are arithmetic means over the log-spaced grid of nine sequence lengths from 128 to 2048.

Inference performance is more GPU dependent. TriHyena is faster than cuEq-TriAttn in 81\% of matched configurations, with a 1.5$\times$ average speedup and a maximum of 5.5$\times$ on GH200 at $L{=}2048$. On H200, it outperforms both baselines at eight of nine sequence lengths; on B300, it outperforms cuEq-TriAttn at eight of nine sequence lengths but is approximately 12\% slower than cuEq-TriMul overall. These runtime gains come with higher memory use: TriHyena has median training peak-memory overheads of 21\% relative to cuEq-TriAttn and 42\% relative to cuEq-TriMul. During inference, it generally uses less memory than cuEq-TriAttn but 20\% more than cuEq-TriMul at the median.




\subsection{TriHyena Scales Efficiently to Wider Pair Representations}
\label{sec:pair-dim-results}
Modern structure models increasingly use wider pair representations, from $d_\mathrm{pair}{=}128$ in AlphaFold2/3 \citep{jumper2021alphafold, abramson2024alphafold3}, Boltz1/2 \citep{wohlwend2024boltz1, passaro2025boltz2}, Protenix-v1 \citep{zhang2026protenixv1}, and RF3 \citep{corley2025rf3}, to $256$ in Protenix-v2 \citep{zhang2026protenixv2} and $384$ in RFD4-Proteina \citep{corley2026rfd4proteina}. We therefore benchmark cuEq-TriAttn, cuEq-TriMul, and TriHyena on NVIDIA B300 and H200 GPUs across pair widths $d_\mathrm{pair}\in{256,384,512,768}$, with $n_\mathrm{head}{=}4$ throughout.

TriHyena’s advantage generally increases with pair width, particularly over cuEq-TriMul (Figures~\ref{fig:b300-training-runtime-channel-benchmark}, \ref{fig:b300-training-memory-channel-benchmark}, \ref{fig:b300-inference-runtime-channel-benchmark}, \ref{fig:b300-inference-memory-channel-benchmark}). During training, TriHyena is fastest in every comparable configuration, averaging $2.3 \times /2.4 \times$ speedups over cuEq-TriAttn/cuEq-TriMul on B300 and $2.5\times /1.9 \times $ on H200. At inference, the corresponding speedups are $1.5\times$/$2.4\times$ on B300 and  $1.9\times$/$2.1\times$ on H200. The only configuration in which cuEq-TriMul is faster than TriHyena occur during inference at $d_\mathrm{pair}$=256 on B300. At $d_\mathrm{pair}{=}768$ and $L{=}1536$, cuEq-TriAttn uses $2.3\times$ and $2.2\times$ more peak training memory than TriHyena on B300 and H200, respectively. At $L{=}2048$, cuEq-TriAttn reaches OOM on both GPUs at $d_\mathrm{pair}$=768, while TriHyena completes successfully.

\subsection{TriHyena Preserves Accuracy in Reduced-scale RF3 Ablations}
\label{sec:rf3-ablation}
To test whether TriHyena can replace TriAttn and/or TriMul without loss of structural accuracy, we ran a controlled, reduced-scale ablation of RF3 \citep{corley2025rf3}. All arms use the full 48-block RF3 Pairformer architecture but are trained under a shortened protocol: 256-token crops, ${\sim}2.55$\,M training examples. We trained three variants of reduced RF3 from scratch under identical conditions, varying only the operators assigned to the four pair-update slots in each of the 48 Pairformer blocks (Table~\ref{tab:rf3-arms}). Each block contains two TriMul and two TriAttn slots, which are replaced in place to preserve the surrounding architecture. Thus, $2\times$ denotes two slots per block and $4\times$ all four; in the $4\times$~TriHyena arm, TriHyena replaces both TriMul and TriAttn entirely. Full training and implementation details are provided in Appendix~\ref{app:rf3-training}.

We evaluate the final checkpoint of each model on a held-out set of 1,206 PDB entries, spanning 32--1,100 tokens and including protein chains, protein--protein interfaces, and protein--ligand interfaces. Each experiment uses 10 seeds and five diffusion samples per seed (Table~\ref{tab:rf3-ablation}). TriHyena filter taps beyond the training offset are zeroed at inference, with unmasked results reported in Appendix~\ref{app:rf3-filter}. Because these are reduced ablation runs, absolute lDDT values support comparisons among arms rather than estimates of full-scale RF3 performance.

\begin{table}[t]
\centering
\caption{RF3 ablation configurations. Pairformer architecture and training conditions are held fixed across arms; only the operators assigned to the TriMul and TriAttn update slots differ.}
\label{tab:rf3-arms}
\small
\begin{tabular}{lccc}
\toprule
Arm & TriMul Slots & TriAttn Slots & Model Parameters \\
\midrule
Baseline (\armBase) & TriMul    & TriAttn   & 356\,M \\
\armHyb             & TriMul    & TriHyena  & 355\,M \\
\armFull            & TriHyena  & TriHyena  & 352\,M \\
\bottomrule
\end{tabular}
\end{table}

\begin{figure}[tbp]
\centering
\includegraphics[width=1.0\linewidth]{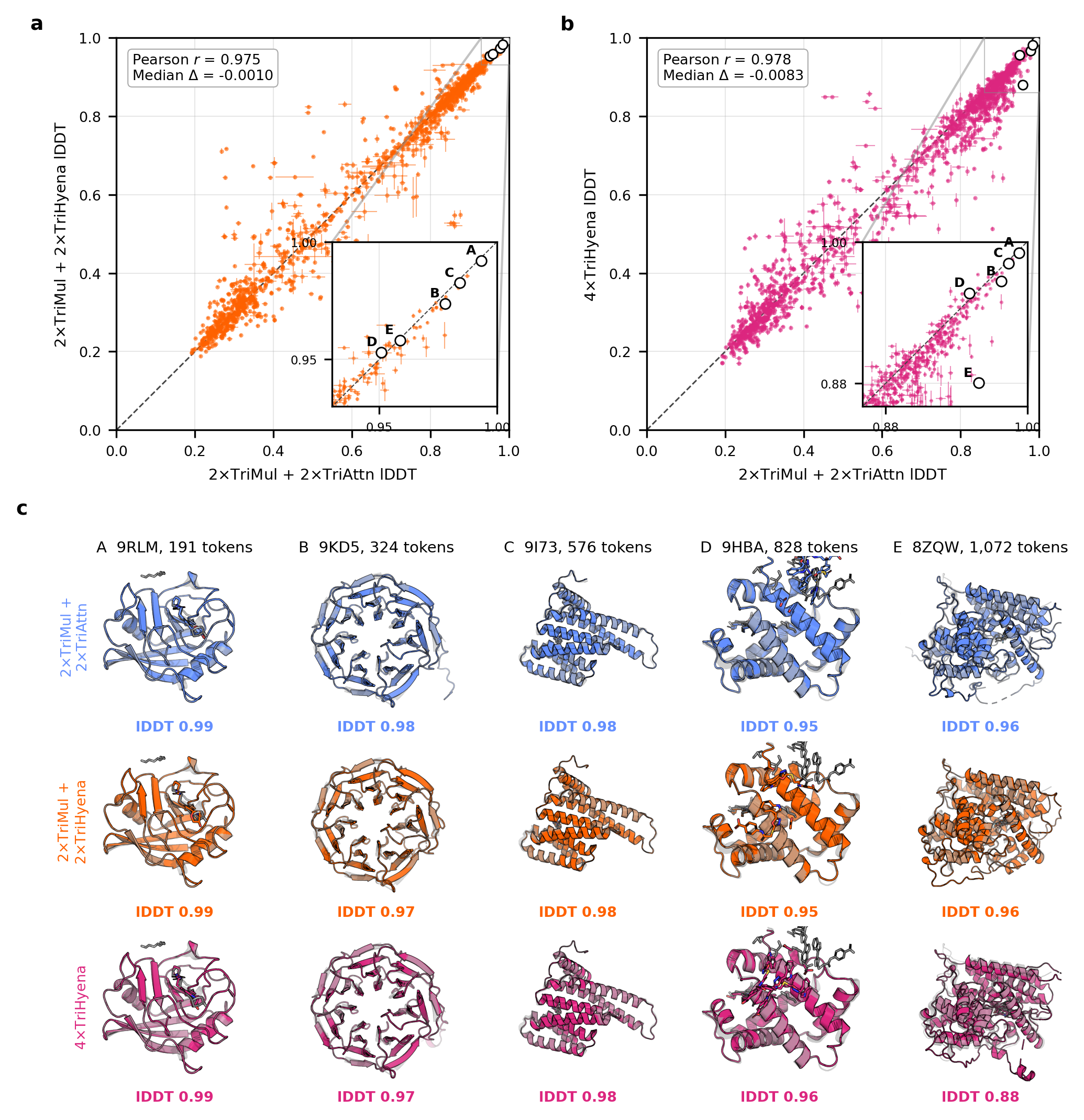}
\caption{TriHyena closely matches the reduced-scale RF3 baseline in per-chain structure prediction accuracy. (a, b) Per-chain protein best-of-1 (BO1) lDDT for the baseline (\armBase) versus (a) \armHyb and (b) \armFull, averaged over 10 seeds. Horizontal and vertical error bars give the standard error of the mean (SEM) over seeds for the baseline and the TriHyena runs, respectively. Boxes report the Pearson correlation and the median per-chain difference (TriHyena $-$ baseline). Insets enlarge the high-lDDT region, and lettered markers denote the entries in (c). (c) Representative predictions spanning 191--1,072 tokens from each arm (colored), superposed on experimental structures (gray). Labels report PDB ID and token count; values below each structure report the lDDT of that run.}
\label{fig:rf3-structures}
\end{figure}

Under this evaluation, replacing TriAttn with TriHyena preserves RF3 accuracy (Table~\ref{tab:rf3-ablation}), matching the baseline on all four metrics. Replacing both TriAttn and TriMul with TriHyena retains near-baseline protein accuracy, within about 0.01 on both protein metrics, but reduces interface lDDT by 0.04 on both protein--protein and protein--ligand interfaces. Figure~\ref{fig:rf3-structures} shows close per-chain agreement with the baseline for both TriHyena arms, and the representative structures are predicted at near-identical accuracy, except for the largest example (8ZQW) under \armFull. Together, these results establish TriHyena as an accurate replacement for TriAttn and support further development toward a fully TriHyena-based Pairformer.

\begin{table}[t] 
\centering 
\caption{RF3 ablation accuracy measured by lDDT on a held-out test set of 1,206 PDB entries. Best-of-1 (BO1) and best-of-5 (BO5) denote the best of one and five diffusion samples, respectively. Values are mean $\pm$ standard deviation over 10 random seeds.} 
\label{tab:rf3-ablation} 
\small 
\begin{tabular}{lcccc} 
\toprule 
Arm & Protein BO1 & Protein BO5 & Protein--Protein BO1 & Protein--Ligand BO1 \\ 
\midrule 
Baseline (\armBase) & 0.641 $\pm$ 0.001 & 0.654 $\pm$ 0.000 & 0.290 $\pm$ 0.003 & 0.425 $\pm$ 0.005 \\ 
\armHyb             & 0.644 $\pm$ 0.001 & 0.655 $\pm$ 0.001 & 0.288 $\pm$ 0.003 & 0.425 $\pm$ 0.005 \\ 
\armFull            & 0.629 $\pm$ 0.001 & 0.644 $\pm$ 0.000 & 0.251 $\pm$ 0.003 & 0.386 $\pm$ 0.005 \\ 
\bottomrule
\end{tabular} 
\end{table}

\subsection{TriHyena Preserves Genie2 Performance} 
\label{sec:genie2-ablation}
To assess whether TriHyena can replace TriMul in generative protein modeling, we trained two variants of Genie2 \citep{lin2024genie2}: the original pair stack containing five TriMul layers and a variant in which all five were replaced by TriHyena. Both models were otherwise trained following the original Genie2 protocol. Full training and evaluation details are provided in Appendix~\ref{app:genie2-training}.

For each evaluated checkpoint, we generated five unconditional backbones at every integer length from 50 to 256 residues, yielding 1,035 structures per checkpoint. Following the Genie2 evaluation protocol, generated backbones were evaluated for designability by sequence design and structure refolding and clustered by structural similarity to quantify diversity. F1 summarizes the resulting designability--diversity trade-off. Full metric definitions and evaluation procedures are provided in Appendix~\ref{app:genie2-evaluation}.

Replacing all five TriMul layers with TriHyena largely preserves Genie2 performance (Table~\ref{tab:Genie2_metrics}). Mean designability, diversity, and F1 are comparable between TriHyena and TriMul, with all differences falling within one standard deviation across checkpoints.

\begin{table}[t] 
\centering 
\caption{Genie2 generative performance for TriMul and TriHyena over the converged checkpoints (epochs 31--40 for TriMul, 31--48 for TriHyena). Designability is the fraction of designable generated structures, and diversity is the fraction of distinct structural clusters among them. F1 score summarizes the designability--diversity trade-off and is averaged over per-epoch F1 scores rather than computed from the mean metrics. Values are mean $\pm$ standard deviation across evaluated checkpoints.} 
\label{tab:Genie2_metrics}
\small 
\begin{tabular}{lcc} 
\toprule 
Metric & TriMul Genie2 & TriHyena Genie2 \\ 
\midrule 
Designability & $0.91 \pm 0.06$ & $0.94 \pm 0.04$ \\ 
Diversity & $0.79 \pm 0.07$ & $0.77 \pm 0.07$ \\ 
F1 score & $0.85 \pm 0.06$ & $0.84 \pm 0.05$ \\ 
\bottomrule 
\end{tabular} 
\end{table}

\section{Discussion}
\label{sec:discussion}


\paragraph{TriHyena shifts the compute--accuracy trade-off.}
TriHyena reduces the computational cost of triangle mixing relative to optimized TriAttn and TriMul, with runtime benefits that generally increase with sequence length and pair width. This scaling is particularly relevant as structure models adopt wider pair representations, where conventional triangle operators become increasingly costly and memory limiting.

\paragraph{TriHyena largely preserves accuracy across triangle mixing operators.}
TriHyena substitutes for TriAttn in reduced-scale RF3 and for TriMul in Genie2 without measurable accuracy loss. Replacing both operators with TriHyena in reduced-scale RF3 reduces interface lDDT by 0.04, suggesting that retaining at least one conventional triangle operator may benefit interface accuracy. However, the current design cannot distinguish this effect from length extrapolation, where the reduced-scale training uses 256-token crops but test sequences extend to 1,100 tokens, or from sensitivity to sequence ordering.

\paragraph{Sequence ordering and causality remain open design choices.}
TriHyena's global filter depends on the sequence-relative offset $(j-k)$ and is therefore sensitive to token ordering. This may contribute to the interface lDDT reduction, since offsets between chains or ligands depend on their input ordering. We have not evaluated sensitivity to reordering or the effects of the causal short convolution and two-sided global filter, making these important directions for further ablation.

\paragraph{Extending TriHyena to longer-range structure modeling.} 
TriHyena offers opportunities for further scaling beyond the sequence lengths explored here. Because the SIREN filter is trained only on offsets within the training crop, longer-range modeling could benefit from learnable channel-wise decay windows that regularize the effective filter range and reduce reliance on manual truncation. Such decay windows are already implemented in nvSubquadratic and, together with training on longer sequences, provide a direct path toward extending TriHyena to longer-range structure modeling.

\section{Acknowledgements}
\label{sec:acknowledgements}

The authors thank Zahra Ronaghi and Farhad Ramezanghorbani for supporting this project. We thank John St. John for insightful discussions on potential extensions of this work. The authors declare no financial interest or conflicts.

\bibliographystyle{unsrtnat}
\bibliography{paper}  

\newpage 

\appendix

\renewcommand{\thefigure}{A\arabic{figure}}
\renewcommand{\thetable}{A\arabic{table}}
\renewcommand{\theequation}{A\arabic{equation}}
\renewcommand{\thealgorithm}{A\arabic{algorithm}}
\renewcommand{\thetheorem}{A\arabic{theorem}}
\setcounter{figure}{0}
\setcounter{table}{0}
\setcounter{equation}{0}
\setcounter{algorithm}{0}
\setcounter{theorem}{0}

\section{$n$-index Generalization}
\label{app:nindex}

This appendix develops the general $n$-index Hyena framework from which TriHyena ($n{=}3$) is instantiated.

\subsection{$n$-index Attention}
\label{app:attn-n}

\begin{definition}[$n$-index attention, $\mathcal{A}_n$]
\label{def:attn-n}
Let $\mathbf{U} \in \mathbb{R}^{B \times L \times \cdots \times L \times d_{\mathrm{pair}}}$, with $n-1$ spatial axes of length $L$, be an $n$-index interaction tensor. Write $\bar{\imath} = (i_1,\ldots,i_{n-1}) \in [L]^{n-1}$ for the saved output cell, $\bar{\imath}' = (i_1,\ldots,i_{n-2})$ for its leading indices, and $i_n$ for the summed index. (For $n{=}3$ in the starting-node orientation, $\bar{\imath}=(i,j)$, $\bar{\imath}'=(i)$ and $i_n=k$, matching Section~\ref{sec:triattn}; for $n{=}2$, $\bar{\imath}=(j)$, $\bar{\imath}'$ is empty and $i_n=k$.) Form a query $\mathbf{Q}_{\bar{\imath}} \in \mathbb{R}^{d_k}$ by applying $W_Q \in \mathbb{R}^{d_{\mathrm{pair}}\times d_k}$ to the feature at cell $\bar{\imath}$, and keys/values $\mathbf{K}_{\bar{\imath}',i_n}, \mathbf{V}^{\mathrm{att}}_{\bar{\imath}',i_n} \in \mathbb{R}^{d_k}$ by applying $W_K, W_V \in \mathbb{R}^{d_{\mathrm{pair}}\times d_k}$ to the feature at the cell $(\bar{\imath}',i_n)$, i.e.\ the output cell with its last index replaced by the summed index. Define:
\begin{equation}
  \mathcal{A}_n(\mathbf{U})_{\bar{\imath}}
  = \sum_{i_n=1}^{L}
    A_{\bar{\imath},\,i_n}(\mathbf{U})\; \mathbf{V}^{\mathrm{att}}_{\bar{\imath}',i_n} W_{\mathrm{out}},
  \qquad
  A_{\bar{\imath},\,i_n}
  = \softmax_{i_n}\!\Bigl(
      \mathbf{Q}_{\bar{\imath}}^{\mathsf{T}} \mathbf{K}_{\bar{\imath}',i_n} / \sqrt{d_k} + b_{\bar{\imath},i_n}
    \Bigr),
  \label{eq:attn-n}
\end{equation}
where $W_{\mathrm{out}} \in \mathbb{R}^{d_k \times d_{\mathrm{pair}}}$ is a learned output projection (for one head; with $n_{\mathrm{head}}$ concatenated heads of width $d_k=d_{\mathrm{hidden}}$ it is $\mathbb{R}^{n_{\mathrm{head}}d_{\mathrm{hidden}}\times d_{\mathrm{pair}}}$, as in Section~\ref{sec:triattn}) and $b_{\bar{\imath},i_n}$ is the attention bias of Eq.~\eqref{eq:triattn} (zero for standard self-attention; for triangle attention it depends only on $(j,k)$). The $n{=}2$ case is standard self-attention and the $n{=}3$ starting-node pattern is triangle attention \citep{jumper2021alphafold}.
\end{definition}

\subsection{$n$-index Hyena Operator}
\label{app:nhyena}

\begin{definition}[$n$-index Hyena Operator, $\mathcal{H}_n$]
\label{def:nhyena}
For $n \geq 2$, let $\mathbf{U} \in \mathbb{R}^{B \times L_1 \times \cdots \times L_{n-1} \times d_{\mathrm{pair}}}$ with spatial axis lengths $L_1,\ldots,L_{n-1} \geq 1$. An $n$-index Hyena operator $\mathcal{H}_n : \mathbf{U} \mapsto \mathbf{Y}$ applies the five-stage skeleton of Section~\ref{sec:hyena2} along an active axis set $\mathcal{S} \subseteq [n-1]$:
\begin{enumerate}[leftmargin=*, itemsep=1pt]
  \item \textbf{Projection and local mix.} Project to $3D$ channels with $W_{\mathrm{proj}}$, then apply a short depthwise finite impulse response (FIR) filter of width $w \ll L_\alpha$ along each $\alpha \in \mathcal{S}$.
  \item \textbf{Channel split.} Split the result into $\mathbf{X}^{(1)}, \mathbf{X}^{(2)}, \mathbf{V}$.
  \item \textbf{First gate.} $\mathbf{Z} = \mathbf{X}^{(1)} \odot \mathbf{V}$.
  \item \textbf{Global mix.} $\mathbf{Z}' = \mathbf{h} \ast_\mathcal{S} \mathbf{Z}$, where $\mathbf{h}$ is a learned $|\mathcal{S}|$-dimensional convolutional kernel; for $\mathcal{S}=\{\alpha\}$, $\ast_\mathcal{S}$ is the single-axis convolution $\ast_\alpha$ of Section~\ref{sec:hyena2}.
  \item \textbf{Second gate and output.} $\mathbf{Y} = (\mathbf{X}^{(2)} \odot \mathbf{Z}')\,W_\mathrm{out}$.
\end{enumerate}
Here, $D$ is the internal channel width of Section~\ref{sec:hyena2} (a hyperparameter not constrained relative to $d_{\mathrm{pair}}$), the filter $\mathbf{h}$ is depthwise (one filter per channel), and $W_{\mathrm{out}}\in\mathbb{R}^{D\times d_{\mathrm{pair}}}$ maps the internal width back to the input channel width $d_{\mathrm{pair}}$ by right-multiplication. When $\mathcal{S}=[n-1]$ and $h_{t_1,\ldots,t_{n-1}}=\prod_{\alpha=1}^{n-1}h^{(\alpha)}_{t_\alpha}$, we call $\mathcal{H}_n$ \emph{separable}. Setting $n{=}2$ recovers standard Hyena \citep{poli2023hyena}. Setting $n{=}3$, $|\mathcal{S}|{=}1$, with $\mathbf{X}^{(1)},\mathbf{X}^{(2)}$ read at the cells $(i,k)$ and $(i,j)$ and the value stream $\mathbf{V}$ read from the transposed cell $(k,i)$ (a permuted index view in the channel split), gives TriHyena, $\mathcal{H}_3$ (Eq.~\eqref{eq:trihyena-output}).
\end{definition}


\subsection{Complexity Analysis}
\label{app:complexity}

Assume uniform axis lengths $L_1 = \cdots = L_{n-1} = L$; set $P = L^{n-1}$. Channel widths $d_{\mathrm{pair}}$ and $D$ and batch size $B$ are constant prefactors, and the interaction order $n$ is treated as a constant in all $O(\cdot)$ and $\Theta(\cdot)$ statements unless an explicit dependence on $n$ is shown.

The cost statements below concern the operation counts of specific algorithms: direct evaluation of the attention sum, and FFT-based evaluation of the Hyena convolution. They are not lower bounds on the underlying computational problems; we do not prove (or claim) that no faster algorithm exists.

\begin{proposition}[Dense $n$-index attention cost]
\label{prop:attn-cost}
Direct evaluation of a dense $n$-index attention block performs $\Theta(L^n)$ operations.
\end{proposition}
\begin{proof}
There are $L^{n-1}$ output cells, and each requires $L$ logit evaluations $\mathbf{Q}_{\bar{\imath}}^{\mathsf{T}}\mathbf{K}_{\bar{\imath}',i_n}$, followed by a softmax and a weighted sum over $i_n$. The total is $\Theta(L^{n-1}\cdot L)=\Theta(L^n)$ for this algorithm.
\end{proof}

\begin{proposition}[Separable $n$-index Hyena cost]
\label{prop:hyena-cost}
FFT-based evaluation of a separable $n$-index Hyena operator with $|\mathcal{S}|=n-1$ active axes performs $\Theta\!\bigl(L^{n-1}(n-1)\log L\bigr)$ operations.
\end{proposition}
\begin{proof}
Following the convention of Section~\ref{sec:hyena2}, stages~1, 3, 5 are pointwise or constant-width local: $O(L^{n-1})$ for constant $w$ and channel widths. Stage~4 decomposes into $n-1$ independent 1D FFT convolutions of length $L$ over the full spatial volume: one sweep along axis $\alpha$ applies $L^{n-2}$ FFT convolutions of length $L$, costing $\Theta(L^{n-2}\cdot L\log L) = \Theta(L^{n-1}\log L)$; summing over $n-1$ axes gives $\Theta(L^{n-1}(n-1)\log L)$, which dominates the pointwise stages.
\end{proof}

\begin{remark}[Single-axis active set]
The same argument applies to any fixed $|\mathcal{S}| \leq n-1$: stage~4 costs $\Theta(|\mathcal{S}|\,L^{n-1}\log L) = \Theta(L^{n-1}\log L)$. In particular, FFT-based TriHyena ($n{=}3$, $|\mathcal{S}|{=}1$) costs $\Theta(L^2\log L)$.
\end{remark}

\begin{remark}[Explicit convolution path]
The bounds above apply to the FFT path. The explicit Tensor Core convolution used on Blackwell GPUs (Section~\ref{sec:implementation}) evaluates each length-$L$ convolution directly in $O(L^2)$ operations, so for TriHyena it costs $O(L^3)$ over all $L$ rows. It trades asymptotic complexity for hardware throughput at short sequence lengths and does not share the $\Theta(L/\log L)$ separation below.
\end{remark}

\begin{corollary}[Complexity separation]
\label{cor:separation}
For any fixed $n \geq 2$, the ratio of the operation count of direct dense $n$-index attention to that of FFT-based separable $n$-index Hyena is $\Theta(L / \log L)$ as $L \to \infty$.
\end{corollary}

\begin{proposition}[Memory complexity] 
\label{prop:memory} 
Separable $n$-index Hyena and memory-efficient dense $n$-index attention that does not materialize the full attention matrix require $O(L^{n-1})$ peak memory. The dominant cost in both cases is the input/output tensor $\mathbf{U} \in \mathbb{R}^{L\times\cdots\times L\times d_{\mathrm{pair}}}$. In contrast, naive attention (e.g. self-attention) that materializes attention weights requires $O(L^n)$ memory. \end{proposition}

\begin{proof} Attention: memory-efficient attention (\emph{e.g.} FlashAttention) computes and consumes the $L$ logits over $i_n$ without materializing the full $L^n$ attention tensor, requiring $O(L)$ working space beyond the $O(L^{n-1})$ input/output buffers. Thus, peak memory is $O(L^{n-1})$. Naive attention (\emph{e.g.} self-attention) instead materializes all $L^n$ attention weights and therefore requires $O(L^n)$ memory. 

Hyena: stages~1, 3, and~5 are pointwise or constant-width local; stage~4 applies 1D FFTs to $L^{n-2}$ fibers, requiring $O(L)$ working space per fiber and $O(L^{n-1})$ total memory. Separable kernel storage is $O(nL)$. \end{proof}

\subsection{Spectral Theory of Separable Kernels}
\label{app:spectral}

\begin{theorem}[Spectral factorization]
\label{thm:factorization}
Let $h_{t_1,\ldots,t_{n-1}} = \prod_{\alpha=1}^{n-1} h^{(\alpha)}_{t_\alpha}$ be a finitely supported separable kernel, with the transform length $L_\alpha$ along each axis taken as the zero-padded length (at least twice the signal length minus one) so that products in frequency space implement linear rather than circular convolution. Its $(n{-}1)$-dimensional DFT factorizes as
\begin{equation}
  \hat{H}(\omega_1,\ldots,\omega_{n-1})
  = \prod_{\alpha=1}^{n-1} \hat{H}^{(\alpha)}(\omega_\alpha),
  \label{eq:factorization}
\end{equation}
where $\hat{H}^{(\alpha)}$ is the 1D discrete Fourier transform (DFT) of $h^{(\alpha)}$. Consequently, the global mix stage of $\mathcal{H}_n$ acts as a single pointwise multiplication by $\hat{H}$ in frequency space, equivalently $n-1$ successive 1D convolutions.
\end{theorem}
\begin{proof}
By definition and separability:
\begin{align*}
  \hat{H}(\omega_1,\ldots,\omega_{n-1})
  &= \sum_{t_1,\ldots,t_{n-1}} \Bigl(\prod_\alpha h^{(\alpha)}_{t_\alpha}\Bigr)
     \prod_\alpha e^{-2\pi \mathrm{j}\, \omega_\alpha t_\alpha / L_\alpha}
  = \prod_{\alpha=1}^{n-1}
     \Bigl(\sum_{t_\alpha} h^{(\alpha)}_{t_\alpha}\,
     e^{-2\pi \mathrm{j}\, \omega_\alpha t_\alpha / L_\alpha}\Bigr)
  = \prod_{\alpha=1}^{n-1} \hat{H}^{(\alpha)}(\omega_\alpha).
\end{align*}
\end{proof}

\paragraph{Exponential-sum parameterization.}
Under $h^{(\alpha)}_\ell = \sum_{r=1}^{R} a_r^{(\alpha)} e^{\lambda_r^{(\alpha)}\ell}$ \citep{gu2022s4} with $\mathrm{Re}\,\lambda_r^{(\alpha)}<0$ (an alternative to the SIREN filters used here), the discrete-time Fourier transform (DTFT) of each factor is rational in $e^{\mathrm{j}\phi}$ with poles at $e^{\lambda_r^{(\alpha)}}$, so $\hat{H}$ is a product of $n{-}1$ rational functions sampled on the unit circle.

\section{Proof Details}
\label{app:proofs}

\subsection{Full Proof of Non-Equivalence (Proposition~\ref{prop:nonequiv})}
\label{app:nonequiv}

For a single channel (Hyena) or head (attention), define the \emph{coupling matrix} of an operator as the $L\times L$ matrix $C$ with $\mathbf{y}_j=\sum_k C_{j,k}\mathbf{v}_k$, where the value stream $\mathbf{v}$ varies and every other quantity (gates, queries, keys, filter) is held fixed.

\begin{proposition}[Non-equivalence with softmax attention]
\label{prop:nonequiv}
Let $n=2$, $L \geq 5$ and $d_k \geq L$, and treat queries and keys as free parameters (equivalently, $d_{\mathrm{pair}}\geq L$ with full-rank $\mathbf{U}$). The set of coupling matrices realizable by $\mathcal{H}_2$ over all gate values and filters has dimension at most $4L-1$, whereas the set realizable by $\mathcal{A}_2$ over all queries and keys has dimension $L(L-1)$. Hence for $L\geq 5$ there are attention coupling matrices that no choice of Hyena gates and filter reproduces, so $\mathcal{H}_2$ and $\mathcal{A}_2$ are not equivalent operator families. The same holds for general $n\geq 2$ by applying the argument to the coupling over one axis with all other indices fixed.
\end{proposition}

\begin{proof}
\noindent\textbf{Hyena coupling.}
Following the convention of Section~\ref{sec:hyena2}, from stages~2--5 the output at position $j$ is $\mathbf{y}_j = \mathbf{x}^{(2)}_{j}\sum_k \mathbf{h}_{j-k}\, \mathbf{x}^{(1)}_{k} \mathbf{v}_k$ (one channel; $\mathbf{W}_{\mathrm{out}}$ omitted). The coupling matrix is $C^{\mathcal{H}}=\mathrm{diag}(\mathbf{x}^{(2)})\,T_{\mathbf{h}}\,\mathrm{diag}(\mathbf{x}^{(1)})$, where $T_{\mathbf{h}}$ is Toeplitz, with entries $C^{\mathcal{H}}_{j,k}=x^{(2)}_j h_{j-k} x^{(1)}_k$. It is the image of the smooth map $(\mathbf{x}^{(2)},\mathbf{h},\mathbf{x}^{(1)})\mapsto C^{\mathcal{H}}$ from $\mathbb{R}^{L}\times\mathbb{R}^{2L-1}\times\mathbb{R}^{L}=\mathbb{R}^{4L-1}$, so, being the image of a smooth map from a lower-dimensional space, its image has Hausdorff dimension at most $4L-1$. (The rescalings $\mathbf{x}^{(2)}\to c\,\mathbf{x}^{(2)}$, $\mathbf{x}^{(1)}\to \mathbf{x}^{(1)}/c$ and $x^{(2)}_j\to\lambda^{-j}x^{(2)}_j$, $h_d\to\lambda^{d}h_d$, $x^{(1)}_k\to\lambda^{k}x^{(1)}_k$ leave $C^{\mathcal{H}}$ unchanged, so the true dimension is slightly smaller, but we only need the upper bound.) The Hyena bound remains valid when the gates are projections of $\mathbf{U}$ rather than free parameters, whereas the attention set is a lower bound only under the free-parameter assumption of the proposition.

\smallskip\noindent\textbf{Attention coupling.}
$C^{\mathcal{A}}_{j,k}=A_{j,k}=\exp(\mathbf{Q}_j^\mathsf{T}\mathbf{K}_k/\sqrt{d_k})/\sum_{k'}\exp(\mathbf{Q}_j^\mathsf{T}\mathbf{K}_{k'}/\sqrt{d_k})$ is row-stochastic. Since $d_k\geq L$, the logit matrix $\Lambda_{j,k}=\mathbf{Q}_j^\mathsf{T}\mathbf{K}_k/\sqrt{d_k}$ can equal any target in $\mathbb{R}^{L\times L}$ via a rank-$L$ factorization; the rowwise softmax then surjects onto the interior of the row-stochastic polytope $\mathcal{R}_L=(\Delta^{L-1})^L$, a set of dimension $L(L-1)$.

\smallskip\noindent\textbf{Dimension gap.}
$4L-1<L(L-1)$ is equivalent to $L^2-5L+1>0$, which holds for all integers $L\geq 5$ (and fails for $L\leq 4$). For such $L$, the Hyena image intersects $\mathcal{R}_L^\circ$ in a set of Hausdorff dimension at most $4L-1<L(L-1)=\dim\mathcal{R}_L$, hence of measure zero in $\mathcal{R}_L$, so almost every strictly positive row-stochastic matrix is an attention coupling that no Hyena gate and filter choice reproduces. For general $n$, fix all spatial indices except one axis pair $(j_\alpha,k_\alpha)$; for each fixed leading index the coupling over that pair has the same form (attention: softmax of query--key logits; Hyena: gated Toeplitz), so the dimension gap applies unchanged. For TriAttn the bias $b_{j,k}$ adds the same kind of $L^2$ degrees of freedom to the logits and does not change the argument. The argument concerns softmax attention (TriAttn); for TriMul the coupling is bilinear in the $(i,k)$ and $(j,k)$ features and lacks Toeplitz structure.

This is a statement about operator families, not about trained models: it rules out exact functional equality, but empirical results \citep{poli2023hyena} show that trained Hyena operators can be competitive despite this structural gap. Accordingly, any parity between TriHyena and TriAttn must be established empirically (Section~\ref{sec:experiments}).
\end{proof}
\subsection{Non-separable Kernels and Parameter Scaling}
\label{app:nonsep}

A fully non-separable $(n{-}1)$-dimensional kernel has $\Theta(L^{n-1})$ entries (and as many parameters if parameterized explicitly; implicit parameterizations such as SIREN decouple the two) and can be applied via $(n{-}1)$-dimensional FFT at cost $\Theta(L^{n-1}(n-1)\log L)$, matching the separable bound. However, storage cost $\Theta(L^{n-1})$ makes non-separable kernels prohibitive for large $n$ or $L$, whereas separable kernels require only $O(nL)$ parameters (one 1D filter per axis).

\section{Implementation}
\label{app:implementation}
Below is pseudocode for the Python implementation of TriHyena. The full implementation is available in the \texttt{nvSubquadratic} repository at \url{https://github.com/NVIDIA-BioNeMo/nvSubquadratic} (accessed 14Oct2026).
\renewcommand{\algorithmiccomment}[1]{\hfill \textcolor{red!70!black}{\textit{#1}}}
\begin{algorithm}[ht]
\caption{TriHyena Forward Pass}
\label{alg:triangle_hyena}
\begin{algorithmic}[1]
\Require $\mathbf{U} \in \mathbb{R}^{B \times L \times L \times d_{\text{pair}}}$,
$\mathbf{m} \in \{0,1\}^{B \times L}$ ($1$ denotes padding),
$\text{start\_node}$ (Boolean),
$D$ (internal channel width)

\State $\mathbf{U} \gets \text{LayerNorm}(\mathbf{U})$

\If{$\text{start\_node}$}
    \State $\mathbf{U}_{\text{row}} \gets
        \text{Reshape}(\mathbf{U},\; [BL,L,d_{\text{pair}}])$
    \State $\mathbf{U}_{\text{col}} \gets
        \text{Reshape}(\mathbf{U}^{\top_{ij}},\; [BL,L,d_{\text{pair}}])$
\Else
    \State $\mathbf{U}_{\text{row}} \gets
        \text{Reshape}(\mathbf{U}^{\top_{ij}},\; [BL,L,d_{\text{pair}}])$
    \State $\mathbf{U}_{\text{col}} \gets
        \text{Reshape}(\mathbf{U},\; [BL,L,d_{\text{pair}}])$
\EndIf

\State $\mathbf{X}^{(12)} \gets
    \text{Transpose}_{1,2}(\text{linear\_x12i}(\mathbf{U}_{\text{row}}))$
    \Comment{$[BL,2D,L]$}
\State $\mathbf{V} \gets
    \text{Transpose}_{1,2}(\text{linear\_vj}(\mathbf{U}_{\text{col}}))$
    \Comment{$[BL,D,L]$}

\State $\mathbf{S} \gets
    \text{Concat}([\mathbf{X}^{(12)},\mathbf{V}],\; \text{dim}=1)$
    \Comment{$[BL,3D,L]$}
\State $\mathbf{S} \gets
    \text{DepthwiseCausalConv1D}(\mathbf{S},\; \text{kernel}=3)$
    \Comment{Preserve length $L$}
\State $\mathbf{X}^{(1)},\mathbf{X}^{(2)},\mathbf{V} \gets
    \text{Split}(\mathbf{S},\; 3,\; \text{dim}=1)$

\State $\mathbf{M}_{\text{pair}}[b,i,j] \gets
    \mathbf{m}[b,i] \lor \mathbf{m}[b,j]$
    \Comment{For all $b,i,j$; shape $[B,L,L]$}
\State $\mathbf{M}_{\text{pad}} \gets
    \text{Reshape}(\mathbf{M}_{\text{pair}},\; [BL,1,L])$
    \Comment{Broadcast across channels}
\State $\mathbf{X}^{(1)} \gets
    \mathbf{X}^{(1)} \odot (1-\mathbf{M}_{\text{pad}})$
\State $\mathbf{X}^{(2)} \gets
    \mathbf{X}^{(2)} \odot (1-\mathbf{M}_{\text{pad}})$

\State $\mathbf{Z} \gets \mathbf{X}^{(1)} \odot \mathbf{V}$
    \Comment{First gate; $[BL,D,L]$}
\State $\mathbf{h} \gets \text{SIRENKernel}(L)$
    \Comment{$[D,2L-1]$; offsets $-(L-1),\ldots,L-1$}
\State $\mathbf{Z} \gets \text{FFTConv1D}(\mathbf{Z},\mathbf{h})$
    \Comment{Two-sided linear convolution; output $[BL,D,L]$}
\State $\mathbf{Z} \gets \mathbf{Z} \odot \mathbf{X}^{(2)}$
    \Comment{Second gate}

\State $\mathbf{Z} \gets \text{Transpose}_{1,2}(\mathbf{Z})$
    \Comment{Restore channels last: $[BL,L,D]$}
\State $\mathbf{Z} \gets \text{Reshape}(\mathbf{Z},\; [B,L,L,D])$
\If{$\lnot\text{start\_node}$}
    \State $\mathbf{Z} \gets \mathbf{Z}^{\top_{ij}}$
\EndIf

\State \Return $\mathbf{Z}W_{\mathrm{out}}+\mathbf{b}_{\mathrm{out}}$
\end{algorithmic}
\end{algorithm}

\newpage

\section{Benchmarking}
\label{app:benchmarking-gpus}

\subsection{GPU benchmark configuration.}

GPU benchmarks were run on NVIDIA B300, B200, GH200, H200, H100, A100, and RTX PRO 6000 Blackwell Server Edition GPUs using CUDA 13.0 and nvSubquadratic-ops 0.3.1. TriHyena, cuEq-TriAttn, and cuEq-TriMul were evaluated in bf16 at sequence lengths $L \in \{128,192,256,384,512,768,1024,1536,2048\}$, with B{=}1, $d_\mathrm{pair}{=}256$, $n_\mathrm{head}{=}4$, $d_\mathrm{hidden}{=}64$, and five replicates per configuration. B300, B200, and RTX PRO 6000 Blackwell Server Edition benchmarks used PyTorch 2.14.1 and cuEquivariance 0.13.0, whereas GH200, H200, H100, and A100 benchmarks used PyTorch 2.13.0 and cuEquivariance 0.11.1. Exact values are shown in Table~\ref{app:gpu_benchmark}.

\begin{figure}[H]
\centering
\includegraphics[width=1.0\linewidth]{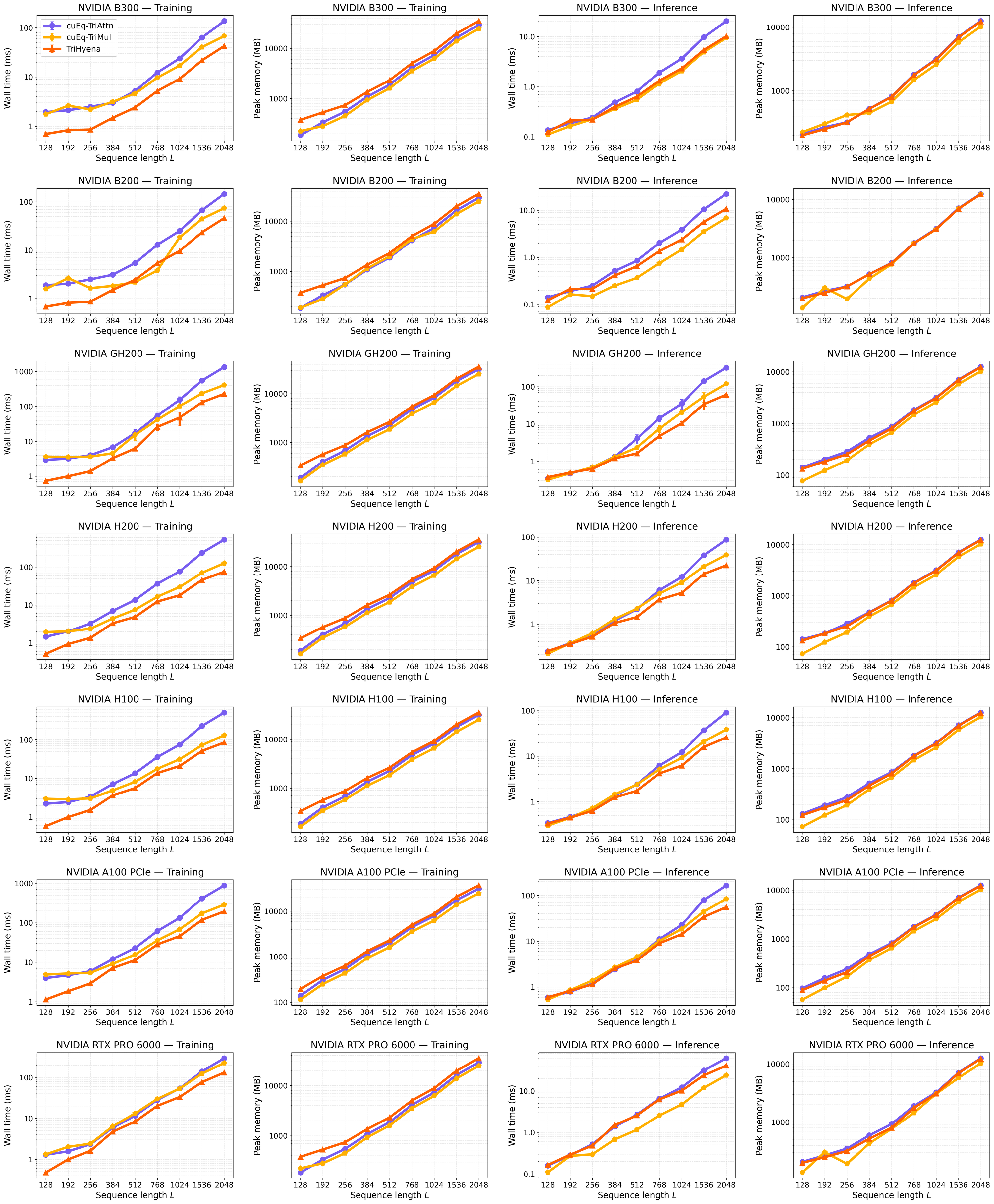}
\caption{Training (forward + backward) and inference (forward only) wall time and peak memory vs. sequence length for cuEq-TriAttn (blue), cuEq-TriMul (gold), and TriHyena (orange) across NVIDIA GPUs (bf16, $B{=}1$, $d_{\mathrm{pair}}{=}256$, $n_{\mathrm{head}}{=}4$; cuEquivariance v0.13.0 on Blackwell and v0.11.1 on Hopper and Ampere; nvSubquadratic-ops v0.3.1 throughout). Points show the mean of five replicates; error bars indicate $\pm 1$ standard deviation.}
\label{fig:nvidia-gpus-benchmark}
\end{figure}

\newpage 

\begingroup
\scriptsize
\setlength{\tabcolsep}{2pt}
\begin{longtable}{llrrrrrrrr}
\toprule
 & & \multicolumn{3}{c}{Wall time (ms)} & \multicolumn{3}{c}{Peak memory (MB)} & \multicolumn{2}{c}{TriHyena speedup} \\
\cmidrule(lr){3-5}\cmidrule(lr){6-8}\cmidrule(lr){9-10}
GPU & $L$ & TriAttn & TriMul & TriHyena & TriAttn & TriMul & TriHyena & vs TriAttn & vs TriMul \\
\midrule
\endfirsthead
\toprule
 & & \multicolumn{3}{c}{Wall time (ms)} & \multicolumn{3}{c}{Peak Memory (MB)} & \multicolumn{2}{c}{TriHyena speedup} \\
\cmidrule(lr){3-5}\cmidrule(lr){6-8}\cmidrule(lr){9-10}
GPU & $L$ & TriAttn & TriMul & TriHyena & TriAttn & TriMul & TriHyena & vs TriAttn & vs TriMul \\
\midrule
\endhead
\midrule
\multicolumn{10}{r}{\emph{continued on next page}} \\
\endfoot
\endlastfoot
\multirow{9}{*}{B300} & 128 & 1.9 & 1.8 & 0.7 & 185 & 225 & 375 & 2.8$\times$ & 2.5$\times$ \\
 & 192 & 2.1 & 2.6 & 0.8 & 336 & 282 & 528 & 2.5$\times$ & 3.1$\times$ \\
 & 256 & 2.5 & 2.2 & 0.9 & 547 & 450 & 736 & 2.9$\times$ & 2.6$\times$ \\
 & 384 & 3.0 & 3.1 & 1.5 & 1,087 & 931 & 1,356 & 2.0$\times$ & 2.1$\times$ \\
 & 512 & 5.2 & 4.6 & 2.4 & 1,883 & 1,605 & 2,309 & 2.2$\times$ & 1.9$\times$ \\
 & 768 & 12.3 & 9.7 & 5.2 & 4,155 & 3,528 & 5,033 & 2.4$\times$ & 1.9$\times$ \\
 & 1024 & 23.8 & 17.0 & 9.1 & 7,337 & 6,222 & 8,845 & 2.6$\times$ & 1.9$\times$ \\
 & 1536 & 63.3 & 40.5 & 21.8 & 16,427 & 13,917 & 19,738 & 2.9$\times$ & 1.9$\times$ \\
 & 2048 & 137 & 68.0 & 42.6 & 29,153 & 24,690 & 34,986 & 3.2$\times$ & 1.6$\times$ \\
\midrule
\multirow{9}{*}{B200} & 128 & 1.9 & 1.6 & 0.7 & 185 & 188 & 375 & 2.8$\times$ & 2.3$\times$ \\
 & 192 & 2.0 & 2.6 & 0.8 & 336 & 282 & 528 & 2.5$\times$ & 3.3$\times$ \\
 & 256 & 2.5 & 1.6 & 0.9 & 547 & 549 & 736 & 2.9$\times$ & 1.9$\times$ \\
 & 384 & 3.1 & 1.8 & 1.5 & 1,087 & 1,150 & 1,356 & 2.1$\times$ & 1.2$\times$ \\
 & 512 & 5.4 & 2.2 & 2.4 & 1,883 & 1,992 & 2,309 & 2.2$\times$ & 0.9$\times$ \\
 & 768 & 13.0 & 3.8 & 5.4 & 4,155 & 4,398 & 5,033 & 2.4$\times$ & 0.7$\times$ \\
 & 1024 & 25.0 & 18.4 & 9.7 & 7,337 & 6,222 & 8,845 & 2.6$\times$ & 1.9$\times$ \\
 & 1536 & 67.0 & 44.4 & 23.5 & 16,427 & 13,917 & 19,738 & 2.9$\times$ & 1.9$\times$ \\
 & 2048 & 146 & 74.2 & 46.2 & 29,153 & 24,690 & 34,986 & 3.2$\times$ & 1.6$\times$ \\
\midrule
\multirow{9}{*}{GH200} & 128 & 2.9 & 3.6 & 0.7 & 187 & 162 & 337 & 4.0$\times$ & 4.9$\times$ \\
 & 192 & 3.2 & 3.6 & 1.0 & 403 & 346 & 568 & 3.2$\times$ & 3.6$\times$ \\
 & 256 & 4.0 & 3.6 & 1.4 & 680 & 578 & 873 & 2.9$\times$ & 2.6$\times$ \\
 & 384 & 6.8 & 4.6 & 3.3 & 1,353 & 1,123 & 1,615 & 2.0$\times$ & 1.4$\times$ \\
 & 512 & 17.1 & 15.3 & 6.2 & 2,269 & 1,860 & 2,635 & 2.7$\times$ & 2.5$\times$ \\
 & 768 & 54.2 & 41.5 & 25.9 & 4,768 & 3,847 & 5,419 & 2.1$\times$ & 1.6$\times$ \\
 & 1024 & 154 & 102 & 47.9 & 8,241 & 6,604 & 9,297 & 3.2$\times$ & 2.1$\times$ \\
 & 1536 & 549 & 236 & 131 & 18,045 & 14,360 & 20,251 & 4.2$\times$ & 1.8$\times$ \\
 & 2048 & 1334 & 409 & 228 & 31,745 & 25,194 & 35,562 & 5.8$\times$ & 1.8$\times$ \\
\midrule
\multirow{9}{*}{H200} & 128 & 1.5 & 1.9 & 0.5 & 187 & 162 & 337 & 2.8$\times$ & 3.7$\times$ \\
 & 192 & 2.0 & 2.0 & 0.9 & 403 & 346 & 568 & 2.2$\times$ & 2.2$\times$ \\
 & 256 & 3.2 & 2.4 & 1.4 & 680 & 578 & 873 & 2.4$\times$ & 1.7$\times$ \\
 & 384 & 7.0 & 4.4 & 3.3 & 1,353 & 1,123 & 1,615 & 2.1$\times$ & 1.3$\times$ \\
 & 512 & 13.6 & 7.5 & 4.8 & 2,269 & 1,860 & 2,635 & 2.8$\times$ & 1.6$\times$ \\
 & 768 & 36.4 & 16.6 & 12.3 & 4,768 & 3,847 & 5,419 & 3.0$\times$ & 1.3$\times$ \\
 & 1024 & 76.6 & 29.9 & 18.1 & 8,241 & 6,604 & 9,297 & 4.2$\times$ & 1.6$\times$ \\
 & 1536 & 237 & 70.3 & 46.0 & 18,045 & 14,360 & 20,251 & 5.2$\times$ & 1.5$\times$ \\
 & 2048 & 527 & 127 & 75.1 & 31,745 & 25,194 & 35,562 & 7.0$\times$ & 1.7$\times$ \\
\midrule
\multirow{9}{*}{H100} & 128 & 2.2 & 3.0 & 0.6 & 187 & 162 & 337 & 3.8$\times$ & 5.1$\times$ \\
 & 192 & 2.4 & 2.9 & 1.0 & 403 & 346 & 568 & 2.4$\times$ & 2.9$\times$ \\
 & 256 & 3.4 & 3.0 & 1.5 & 680 & 578 & 873 & 2.2$\times$ & 2.0$\times$ \\
 & 384 & 7.1 & 4.9 & 3.6 & 1,353 & 1,123 & 1,615 & 2.0$\times$ & 1.3$\times$ \\
 & 512 & 13.5 & 8.1 & 5.6 & 2,269 & 1,860 & 2,635 & 2.4$\times$ & 1.5$\times$ \\
 & 768 & 35.5 & 17.6 & 13.7 & 4,768 & 3,847 & 5,419 & 2.6$\times$ & 1.3$\times$ \\
 & 1024 & 74.7 & 31.2 & 20.8 & 8,241 & 6,604 & 9,297 & 3.6$\times$ & 1.5$\times$ \\
 & 1536 & 227 & 72.4 & 51.3 & 18,045 & 14,360 & 20,251 & 4.4$\times$ & 1.4$\times$ \\
 & 2048 & 506 & 130 & 84.8 & 31,745 & 25,194 & 35,562 & 6.0$\times$ & 1.5$\times$ \\
\midrule
\multirow{9}{*}{A100} & 128 & 4.0 & 4.9 & 1.1 & 139 & 114 & 195 & 3.5$\times$ & 4.3$\times$ \\
 & 192 & 4.7 & 5.2 & 1.9 & 307 & 250 & 376 & 2.5$\times$ & 2.8$\times$ \\
 & 256 & 5.9 & 5.4 & 2.9 & 537 & 435 & 631 & 2.0$\times$ & 1.9$\times$ \\
 & 384 & 12.0 & 9.1 & 7.1 & 1,162 & 932 & 1,327 & 1.7$\times$ & 1.3$\times$ \\
 & 512 & 22.9 & 15.6 & 11.3 & 2,030 & 1,621 & 2,297 & 2.0$\times$ & 1.4$\times$ \\
 & 768 & 61.6 & 35.7 & 28.3 & 4,481 & 3,561 & 5,035 & 2.2$\times$ & 1.3$\times$ \\
 & 1024 & 133 & 68.4 & 45.7 & 7,907 & 6,269 & 8,864 & 2.9$\times$ & 1.5$\times$ \\
 & 1536 & 411 & 173 & 118 & 17,663 & 13,978 & 20,922 & 3.5$\times$ & 1.5$\times$ \\
 & 2048 & 879 & 288 & 193 & 31,315 & 24,764 & 37,081 & 4.5$\times$ & 1.5$\times$ \\
\midrule
\multirow{9}{*}{RTX PRO 6000 Blackwell Server Edition} & 128 & 1.3 & 1.3 & 0.5 & 185 & 225 & 381 & 2.7$\times$ & 2.8$\times$ \\
 & 192 & 1.6 & 2.0 & 1.0 & 336 & 282 & 530 & 1.6$\times$ & 2.0$\times$ \\
 & 256 & 2.3 & 2.4 & 1.6 & 547 & 450 & 742 & 1.4$\times$ & 1.5$\times$ \\
 & 384 & 5.9 & 6.4 & 4.7 & 1,077 & 931 & 1,359 & 1.2$\times$ & 1.4$\times$ \\
 & 512 & 11.6 & 13.2 & 8.3 & 1,865 & 1,605 & 2,315 & 1.4$\times$ & 1.6$\times$ \\
 & 768 & 28.2 & 29.9 & 20.2 & 4,115 & 3,528 & 5,035 & 1.4$\times$ & 1.5$\times$ \\
 & 1024 & 54.1 & 53.5 & 33.3 & 7,265 & 6,222 & 8,849 & 1.6$\times$ & 1.6$\times$ \\
 & 1536 & 141 & 124 & 77.0 & 16,265 & 13,917 & 19,739 & 1.8$\times$ & 1.6$\times$ \\
 & 2048 & 294 & 226 & 132 & 28,865 & 24,690 & 34,986 & 2.2$\times$ & 1.7$\times$ \\
\bottomrule
\caption{Training-step (forward\,+\,backward) wall time and peak memory for cuEq-TriAttn, cuEq-TriMul, and TriHyena at $d_{\mathrm{pair}}{=}256$ and $d_{\mathrm{hidden}}{=}64$ (bf16, $B{=}1$, $n_{\mathrm{head}}{=}4$; mean of 5 replicates). Speedup is the baseline wall time divided by the TriHyena wall time ($>1$ favors TriHyena). Peak memory includes one-time initialization allocations.}
\end{longtable}
\label{app:gpu_benchmark}
\endgroup

\newpage 

\subsection{Pair dimension benchmark configuration.}

Pair dimension benchmarks were run on NVIDIA B300 and H200 GPUs using PyTorch 2.14.1 and cuEquivariance 0.13.0 on B300, and PyTorch 2.13.0 and cuEquivariance 0.11.1 on H200. TriHyena, cuEq-TriAttn, and cuEq-TriMul were evaluated in bf16 at sequence lengths $L \in \{256,512,1024,2048\}$, with batch size 1, $n_\mathrm{head}=4$, and five replicates per configuration. Pair dimensions were varied according to $d_\mathrm{pair}=n_\mathrm{head} d_\mathrm{hidden}$ and $(d_\mathrm{pair},d_\mathrm{hidden}) \in \{(384,96),(512,128),(768,192)\}$. Exact values are shown in Table~\ref{tab:dpair-benchmarks}

\begin{figure}[H]
\centering
\includegraphics[width=1.0\linewidth]{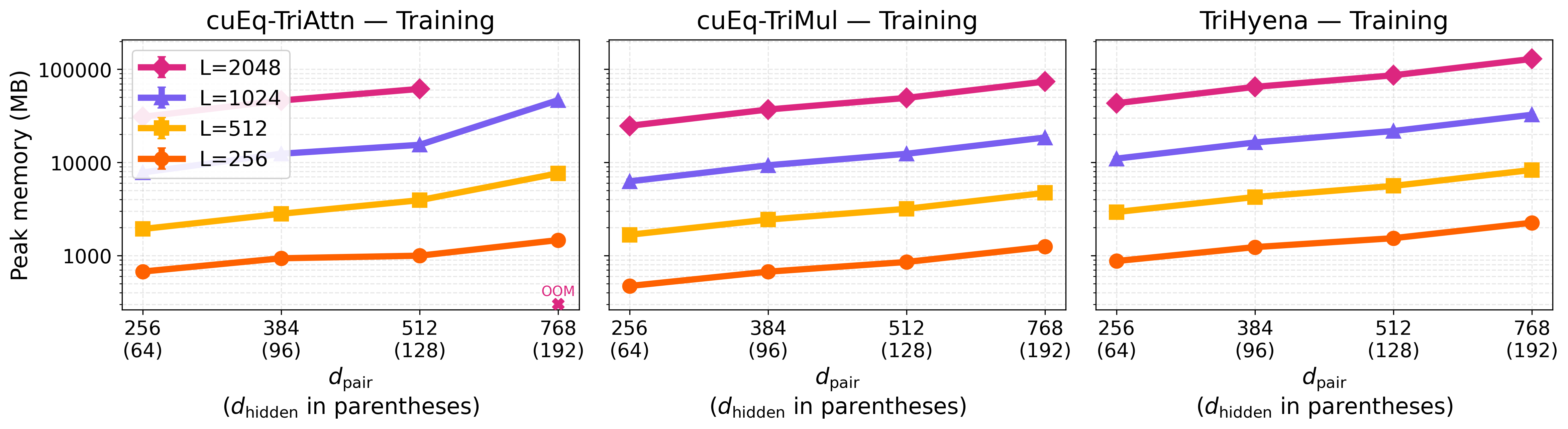}
\caption{Training (forward + backward) peak memory usage vs. pair dimension ($d_{\mathrm{pair}}$) for cuEq-TriAttn, cuEq-TriMul, and TriHyena on an NVIDIA B300 GPU (bf16, $B{=}1$, $n_{\mathrm{head}}{=}4$; cuEquivariance v0.13.0, nvSubquadratic-ops v0.3.1). Colors denote sequence length ($L = 256, 512, 1024, 2048$); $\boldsymbol{\times}$ marks OOM. Points show the mean of five replicates; error bars indicate $\pm 1$ standard deviation.}
\label{fig:b300-training-memory-channel-benchmark}
\end{figure}

\begin{figure}[H]
\centering
\includegraphics[width=1.0\linewidth]{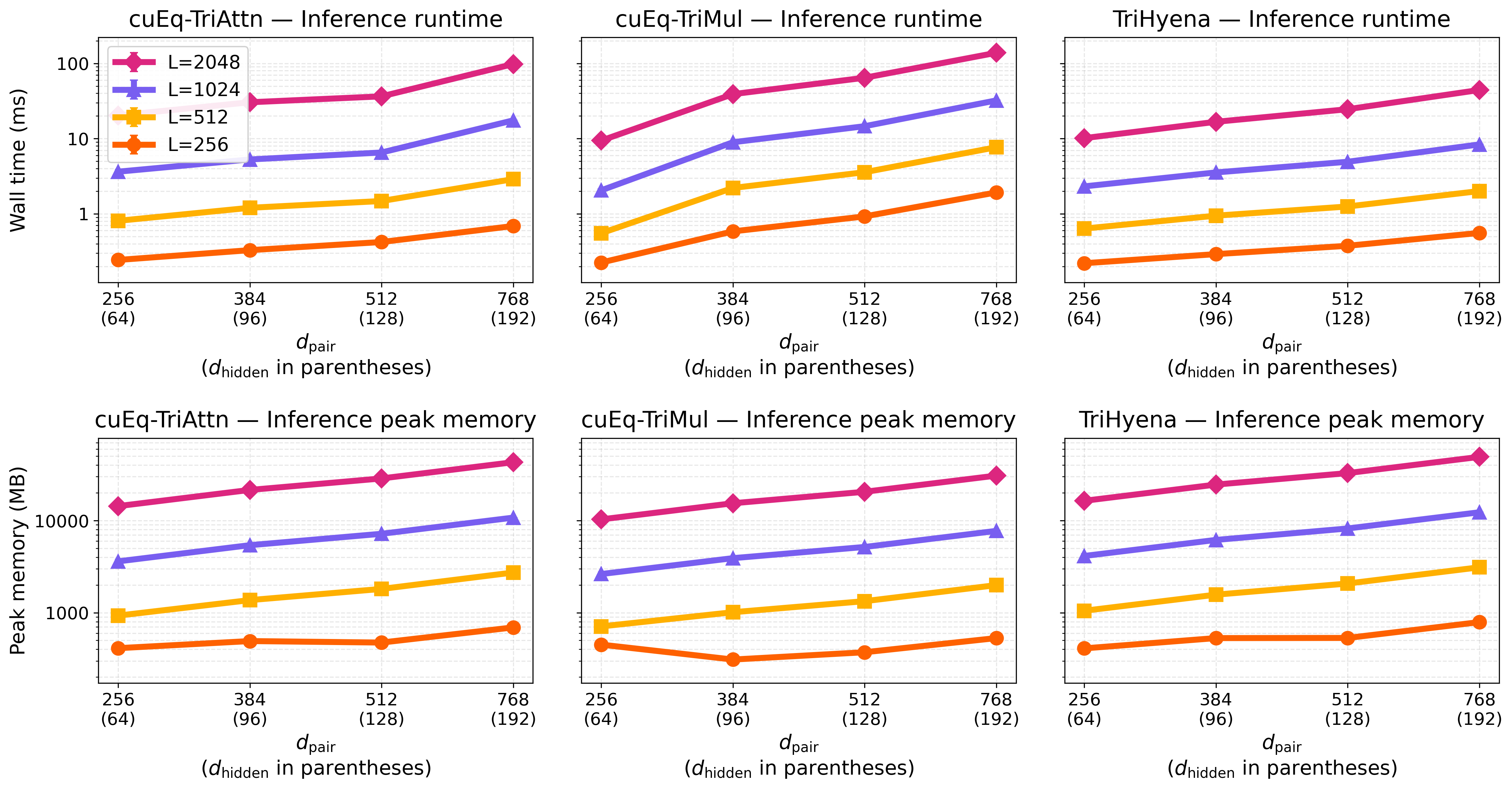}
\caption{Inference (forward only) wall time and peak memory usage vs. pair dimension ($d_{\mathrm{pair}}$) for cuEq-TriAttn, cuEq-TriMul, and TriHyena on an NVIDIA B300 GPU (bf16, $B{=}1$, $n_{\mathrm{head}}{=}4$; cuEquivariance v0.13.0, nvSubquadratic-ops v0.3.1). Colors denote sequence length ($L = 256, 512, 1024, 2048$). Points show the mean of five replicates; error bars indicate $\pm 1$ standard deviation.}
\label{fig:b300-inference-runtime-channel-benchmark}
\end{figure}

\begin{figure}[H]
\centering
\includegraphics[width=1.0\linewidth]{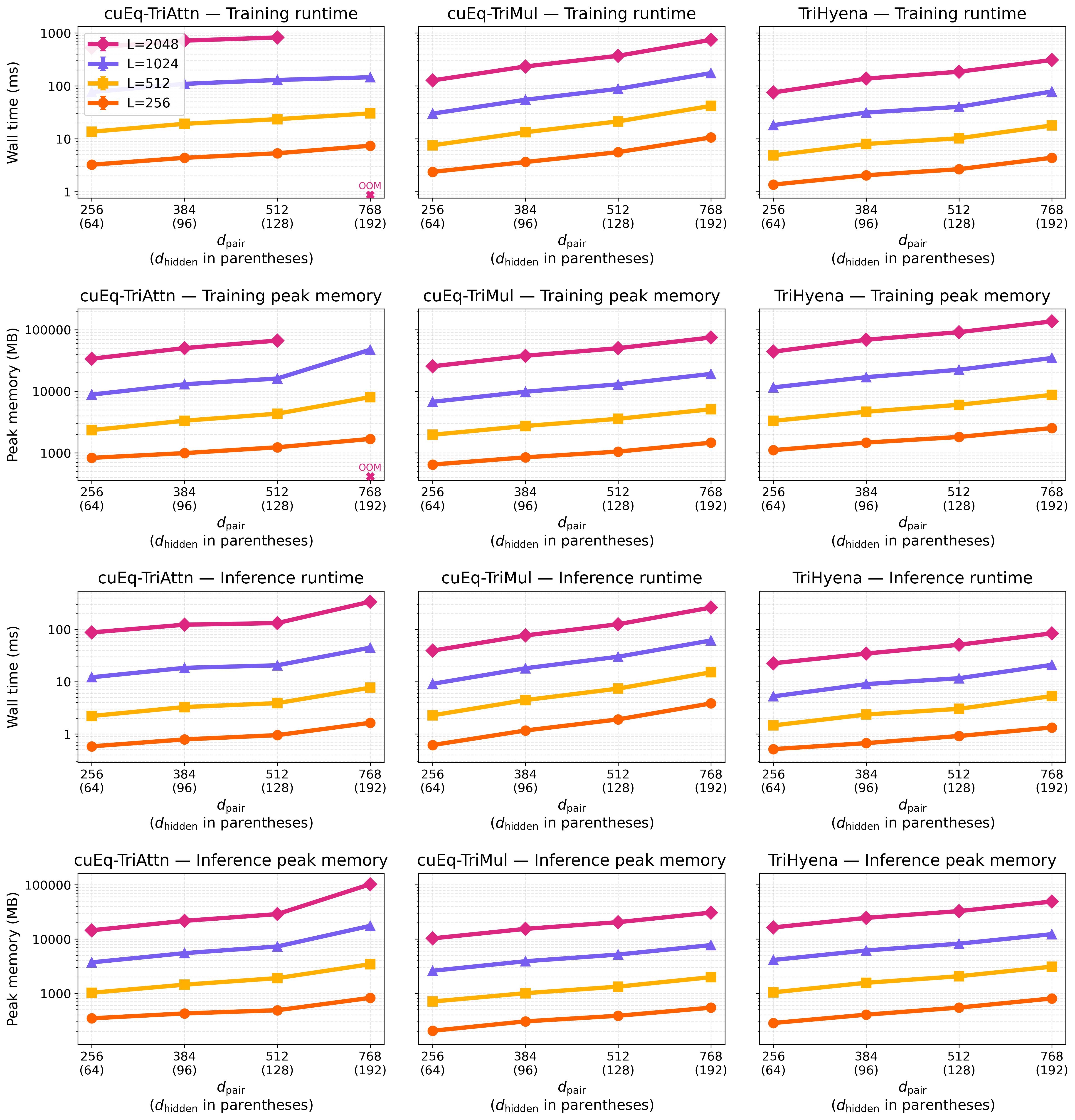}
\caption{Training (forward + backward) and Inference (forward only) wall time and peak memory usage vs. pair dimension ($d_{\mathrm{pair}}$) for cuEq-TriAttn, cuEq-TriMul, and TriHyena on an NVIDIA H200 GPU (bf16, $B{=}1$, $n_{\mathrm{head}}{=}4$; cuEquivariance v0.11.1, nvSubquadratic-ops v0.3.1). Colors denote sequence length ($L = 256, 512, 1024, 2048$); $\boldsymbol{\times}$ marks OOM. Points show the mean of five replicates; error bars indicate $\pm 1$ standard deviation.}
\label{fig:b300-inference-memory-channel-benchmark}
\end{figure}

\newpage 

\begingroup
\footnotesize
\setlength{\tabcolsep}{2pt}
\begin{longtable}{llrrrrrrrr}
\toprule
 & & \multicolumn{3}{c}{Wall time (ms)} & \multicolumn{3}{c}{Peak memory (MB)} & \multicolumn{2}{c}{TriHyena speedup} \\
\cmidrule(lr){3-5}\cmidrule(lr){6-8}\cmidrule(lr){9-10}
$d_{\mathrm{pair}}$ & $L$ & TriAttn & TriMul & TriHyena & TriAttn & TriMul & TriHyena & vs TriAttn & vs TriMul \\
\midrule
\endfirsthead
\toprule
 & & \multicolumn{3}{c}{Wall time (ms)} & \multicolumn{3}{c}{Peak memory (MB)} & \multicolumn{2}{c}{TriHyena speedup} \\
\cmidrule(lr){3-5}\cmidrule(lr){6-8}\cmidrule(lr){9-10}
$d_{\mathrm{pair}}$ & $L$ & TriAttn & TriMul & TriHyena & TriAttn & TriMul & TriHyena & vs TriAttn & vs TriMul \\
\midrule
\endhead
\midrule
\multicolumn{10}{r}{\emph{continued on next page}} \\
\endfoot
\endlastfoot
\multicolumn{10}{l}{\textbf{NVIDIA B300}} \\
\midrule
\multirow{9}{*}{256} & 128 & 1.9 & 1.8 & 0.7 & 185 & 225 & 375 & 2.8$\times$ & 2.5$\times$ \\
 & 192 & 2.1 & 2.6 & 0.8 & 336 & 282 & 528 & 2.5$\times$ & 3.1$\times$ \\
 & 256 & 2.5 & 2.2 & 0.9 & 547 & 450 & 736 & 2.9$\times$ & 2.6$\times$ \\
 & 384 & 3.0 & 3.1 & 1.5 & 1,087 & 931 & 1,356 & 2.0$\times$ & 2.1$\times$ \\
 & 512 & 5.2 & 4.6 & 2.4 & 1,883 & 1,605 & 2,309 & 2.2$\times$ & 1.9$\times$ \\
 & 768 & 12.3 & 9.7 & 5.2 & 4,155 & 3,528 & 5,033 & 2.4$\times$ & 1.9$\times$ \\
 & 1024 & 23.8 & 17.0 & 9.1 & 7,337 & 6,222 & 8,845 & 2.6$\times$ & 1.9$\times$ \\
 & 1536 & 63.3 & 40.5 & 21.8 & 16,427 & 13,917 & 19,738 & 2.9$\times$ & 1.9$\times$ \\
 & 2048 & 137 & 68.0 & 42.6 & 29,153 & 24,690 & 34,986 & 3.2$\times$ & 1.6$\times$ \\
\midrule
\multirow{9}{*}{384} & 128 & 1.9 & 2.5 & 0.8 & 246 & 212 & 436 & 2.6$\times$ & 3.2$\times$ \\
 & 192 & 2.3 & 2.6 & 1.0 & 472 & 392 & 664 & 2.2$\times$ & 2.6$\times$ \\
 & 256 & 2.8 & 2.8 & 1.1 & 744 & 644 & 977 & 2.5$\times$ & 2.5$\times$ \\
 & 384 & 4.1 & 4.7 & 2.2 & 1,592 & 1,365 & 1,969 & 1.9$\times$ & 2.1$\times$ \\
 & 512 & 7.2 & 7.8 & 3.8 & 2,780 & 2,374 & 3,399 & 1.9$\times$ & 2.1$\times$ \\
 & 768 & 17.1 & 17.2 & 7.9 & 6,172 & 5,258 & 7,483 & 2.1$\times$ & 2.2$\times$ \\
 & 1024 & 32.3 & 31.4 & 14.2 & 10,922 & 9,295 & 13,200 & 2.3$\times$ & 2.2$\times$ \\
 & 1536 & 84.9 & 73.2 & 35.4 & 24,492 & 20,830 & 29,533 & 2.4$\times$ & 2.1$\times$ \\
 & 2048 & 177 & 124 & 69.3 & 43,490 & 36,979 & 52,399 & 2.6$\times$ & 1.8$\times$ \\
\midrule
\multirow{9}{*}{512} & 128 & 1.9 & 2.5 & 0.7 & 308 & 262 & 498 & 2.6$\times$ & 3.4$\times$ \\
 & 192 & 2.3 & 2.6 & 1.2 & 608 & 502 & 800 & 1.9$\times$ & 2.2$\times$ \\
 & 256 & 2.9 & 3.3 & 1.3 & 970 & 839 & 1,221 & 2.2$\times$ & 2.5$\times$ \\
 & 384 & 5.0 & 6.8 & 2.9 & 2,098 & 1,800 & 2,583 & 1.7$\times$ & 2.4$\times$ \\
 & 512 & 8.8 & 11.7 & 4.9 & 3,677 & 3,145 & 4,489 & 1.8$\times$ & 2.4$\times$ \\
 & 768 & 20.8 & 26.7 & 10.9 & 8,190 & 6,989 & 9,934 & 1.9$\times$ & 2.5$\times$ \\
 & 1024 & 39.3 & 48.5 & 20.0 & 14,507 & 12,370 & 17,555 & 2.0$\times$ & 2.4$\times$ \\
 & 1536 & 100 & 105 & 50.5 & 32,557 & 27,745 & 39,330 & 2.0$\times$ & 2.1$\times$ \\
 & 2048 & 205 & 198 & 99.4 & 57,827 & 49,270 & 69,812 & 2.1$\times$ & 2.0$\times$ \\
\midrule
\multirow{9}{*}{768} & 128 & 1.8 & 2.6 & 0.8 & 423 & 366 & 621 & 2.2$\times$ & 3.1$\times$ \\
 & 192 & 3.0 & 3.5 & 1.8 & 826 & 726 & 1,073 & 1.7$\times$ & 2.0$\times$ \\
 & 256 & 5.2 & 6.0 & 2.1 & 1,446 & 1,230 & 1,769 & 2.5$\times$ & 2.9$\times$ \\
 & 384 & 10.8 & 13.1 & 4.5 & 3,312 & 2,671 & 3,812 & 2.4$\times$ & 2.9$\times$ \\
 & 512 & 19.8 & 23.4 & 7.9 & 6,857 & 4,689 & 6,670 & 2.5$\times$ & 3.0$\times$ \\
 & 768 & 51.1 & 53.6 & 18.4 & 19,947 & 10,452 & 14,836 & 2.8$\times$ & 2.9$\times$ \\
 & 1024 & 101 & 98.9 & 34.5 & 43,599 & 18,522 & 26,266 & 2.9$\times$ & 2.9$\times$ \\
 & 1536 & 286 & 215 & 88.9 & 134,873 & 41,577 & 58,923 & 3.2$\times$ & 2.4$\times$ \\
 & 2048 & \textsc{oom} & 400 & 175 & \textsc{oom} & 73,854 & 104,639 & \textsc{oom} & 2.3$\times$ \\
\midrule
\multicolumn{10}{l}{\textbf{NVIDIA H200}} \\
\midrule
\multirow{9}{*}{256} & 128 & 1.5 & 1.9 & 0.5 & 187 & 162 & 337 & 2.8$\times$ & 3.7$\times$ \\
 & 192 & 2.0 & 2.0 & 0.9 & 403 & 346 & 568 & 2.2$\times$ & 2.2$\times$ \\
 & 256 & 3.2 & 2.4 & 1.4 & 680 & 578 & 873 & 2.4$\times$ & 1.7$\times$ \\
 & 384 & 7.0 & 4.4 & 3.3 & 1,353 & 1,123 & 1,615 & 2.1$\times$ & 1.3$\times$ \\
 & 512 & 13.6 & 7.5 & 4.8 & 2,269 & 1,860 & 2,635 & 2.8$\times$ & 1.6$\times$ \\
 & 768 & 36.4 & 16.6 & 12.3 & 4,768 & 3,847 & 5,419 & 3.0$\times$ & 1.3$\times$ \\
 & 1024 & 76.6 & 29.9 & 18.1 & 8,241 & 6,604 & 9,297 & 4.2$\times$ & 1.6$\times$ \\
 & 1536 & 237 & 70.3 & 46.0 & 18,045 & 14,360 & 20,251 & 5.2$\times$ & 1.5$\times$ \\
 & 2048 & 527 & 127 & 75.1 & 31,745 & 25,194 & 35,562 & 7.0$\times$ & 1.7$\times$ \\
\midrule
\multirow{9}{*}{384} & 128 & 1.6 & 1.9 & 0.7 & 247 & 212 & 404 & 2.3$\times$ & 2.7$\times$ \\
 & 192 & 2.5 & 2.3 & 1.5 & 539 & 456 & 721 & 1.7$\times$ & 1.6$\times$ \\
 & 256 & 4.4 & 3.6 & 2.0 & 921 & 772 & 1,144 & 2.1$\times$ & 1.8$\times$ \\
 & 384 & 9.9 & 7.6 & 5.6 & 1,893 & 1,557 & 2,227 & 1.8$\times$ & 1.4$\times$ \\
 & 512 & 19.2 & 13.3 & 8.0 & 3,230 & 2,630 & 3,722 & 2.4$\times$ & 1.7$\times$ \\
 & 768 & 51.4 & 30.1 & 19.4 & 6,929 & 5,577 & 7,868 & 2.6$\times$ & 1.6$\times$ \\
 & 1024 & 109 & 54.6 & 31.3 & 12,082 & 9,677 & 13,648 & 3.5$\times$ & 1.8$\times$ \\
 & 1536 & 325 & 129 & 83.1 & 26,686 & 21,274 & 31,770 & 3.9$\times$ & 1.6$\times$ \\
 & 2048 & 716 & 232 & 137 & 47,106 & 37,483 & 56,041 & 5.2$\times$ & 1.7$\times$ \\
\midrule
\multirow{9}{*}{512} & 128 & 1.7 & 2.0 & 0.9 & 309 & 262 & 475 & 2.0$\times$ & 2.4$\times$ \\
 & 192 & 3.1 & 3.3 & 1.8 & 675 & 566 & 876 & 1.7$\times$ & 1.9$\times$ \\
 & 256 & 5.3 & 5.6 & 2.7 & 1,162 & 967 & 1,419 & 2.0$\times$ & 2.1$\times$ \\
 & 384 & 12.2 & 12.1 & 6.8 & 2,435 & 1,991 & 2,841 & 1.8$\times$ & 1.8$\times$ \\
 & 512 & 23.5 & 21.3 & 10.2 & 4,191 & 3,401 & 4,813 & 2.3$\times$ & 2.1$\times$ \\
 & 768 & 62.7 & 48.6 & 24.2 & 9,090 & 7,308 & 10,319 & 2.6$\times$ & 2.0$\times$ \\
 & 1024 & 129 & 87.9 & 40.1 & 15,923 & 12,752 & 18,003 & 3.2$\times$ & 2.2$\times$ \\
 & 1536 & 379 & 200 & 111 & 35,327 & 28,189 & 42,141 & 3.4$\times$ & 1.8$\times$ \\
 & 2048 & 822 & 370 & 185 & 62,467 & 49,774 & 74,476 & 4.5$\times$ & 2.0$\times$ \\
\midrule
\multirow{9}{*}{768} & 128 & 2.0 & 2.7 & 1.3 & 423 & 366 & 612 & 1.5$\times$ & 2.0$\times$ \\
 & 192 & 4.3 & 6.2 & 3.0 & 890 & 790 & 1,184 & 1.4$\times$ & 2.1$\times$ \\
 & 256 & 7.4 & 10.6 & 4.4 & 1,574 & 1,358 & 1,964 & 1.7$\times$ & 2.4$\times$ \\
 & 384 & 16.5 & 23.5 & 11.0 & 3,504 & 2,863 & 4,068 & 1.5$\times$ & 2.1$\times$ \\
 & 512 & 30.2 & 41.8 & 17.9 & 7,113 & 4,944 & 6,990 & 1.7$\times$ & 2.3$\times$ \\
 & 768 & 75.5 & 97.1 & 47.3 & 20,267 & 10,771 & 16,080 & 1.6$\times$ & 2.0$\times$ \\
 & 1024 & 145 & 174 & 78.2 & 43,983 & 18,904 & 28,244 & 1.9$\times$ & 2.2$\times$ \\
 & 1536 & 500 & 400 & 182 & 135,321 & 42,020 & 62,883 & 2.8$\times$ & 2.2$\times$ \\
 & 2048 & \textsc{oom} & 741 & 308 & \textsc{oom} & 74,358 & 111,347 & \textsc{oom} & 2.4$\times$ \\
\bottomrule
\caption{Training-step (forward\,+\,backward) wall time and peak memory for cuEq-TriAttn, cuEq-TriMul, and TriHyena on NVIDIA B300 and H200 GPUs (bf16, $B{=}1$, $n_{\mathrm{head}}{=}4$, $d_{\mathrm{pair}}{=}n_{\mathrm{head}}d_{\mathrm{hidden}}$; mean of 5 replicates). Speedup is baseline wall time divided by TriHyena wall time ($>1$ favors TriHyena). OOM denotes an out-of-memory failure; peak memory includes one-time initialization allocations.}
\label{tab:dpair-benchmarks}
\end{longtable}
\endgroup


\section{RF3 Ablation Training and Evaluation Details}
\label{app:rf3-training}

RF3 \citep{corley2025rf3} ablation experiments were conducted on NVIDIA H200 GPUs using bfloat16 (bf16) mixed precision with PyTorch 2.10.0a0, CUDA 13.0, cuDNN 9.15, and NCCL 2.28. TriHyena used nvSubquadratic-ops v0.2.1 while cuEq-TriAttn and cuEq-TriMul used cuEquivariance 0.9.1. 


\paragraph{Architecture.}
Each RF3 pairformer block contains four residual \texttt{(pair, is\_padding)} $\rightarrow$ \texttt{pair} updates in the fixed order \texttt{tri\_mul\_outgoing}, \texttt{tri\_mul\_incoming}, \texttt{tri\_attn\_start}, and \texttt{tri\_attn\_end}. Across the three ablation arms (Table~\ref{tab:rf3-arms}), these slots were replaced rather than removed. The 48-block architecture, slot order, residual wiring, dropout axes, transitions, single-sequence track, and $d_{\mathrm{pair}}=128$ were identical across arms. Each TriHyena instance used internal width $D=128=d_{\mathrm{pair}}$.

\paragraph{Training protocol.}
All three models were trained from random initialization on $32\times$NVIDIA H200 GPUs (4 nodes $\times$ 8 GPUs) using DDP and bf16 mixed precision. Training used an effective batch size of 64 with gradient accumulation of 2, a 256-token crop (maximum 2,176 atoms), four recycles, and a diffusion batch of 48 per example. We used Adam with $\beta=(0.9,0.95)$ and $\epsilon=10^{-8}$, the AlphaFold3 learning-rate schedule \citep{abramson2024alphafold3}, a base learning rate of $1.8\times10^{-3}$, 1,000 warm-up steps, a decay factor of $0.95$ every 50,000 steps, and gradient-norm clipping at $10.0$.

The training mixture comprised PDB ($p=0.41$) and an RF3-predicted MGnify distillation set \citep{richardson2023mgnify} ($p=0.59$), including the nucleic-acid, RNA-monomer, and inverse-folding distillation subsets. Each run covered 106 epochs of 24,000 examples (39,750 optimizer steps; ${\sim}2.55$\,M examples) using a shared random seed. Validation was performed every 10 epochs on the AlphaFold3 validation split restricted to targets with fewer than 512 tokens.


The baseline contains $0.9$\,M more parameters than the \armHyb arm ($+0.26\%$) and $3.3$\,M more than the \armFull arm ($+0.94\%$); the comparison therefore controls architecture and training protocol but not parameter count exactly.

\paragraph{Evaluation protocol.}
We evaluate the final checkpoint (epoch 105) on a held-out test set of 1,206 PDB entries, spanning 32--1,100 tokens. The test set contains 1,506 protein chains, 400 protein--protein interfaces, and 399 protein--ligand interfaces. Each experiment uses 10 recycles and five diffusion samples per seed. Table~\ref{tab:rf3-ablation} reports mean $\pm$ standard deviation across seeds for protein best-of-1 (BO1), protein best-of-5 (BO5), protein--protein BO1, and protein--ligand BO1 lDDT.

\paragraph{TriHyena filter handling at inference.}
Training uses 256-token crops ($L_{\mathrm{train}}=256$), so the TriHyena SIREN filter is trained only on offsets $|j-k|\leq L_{\mathrm{train}}-1=255$. At inference, filter taps beyond this range are set to zero. This leaves TriHyena unchanged for sequences of up to $L_{\mathrm{train}}$ tokens while permitting evaluation on the longer sequences in the test set without retraining. Results without this masking are reported in Appendix~\ref{app:rf3-filter}.


\section{RF3: TriHyena filter beyond the training crop}
\label{app:rf3-filter}

\begin{figure}[H]
\centering
\includegraphics[width=0.6\linewidth]{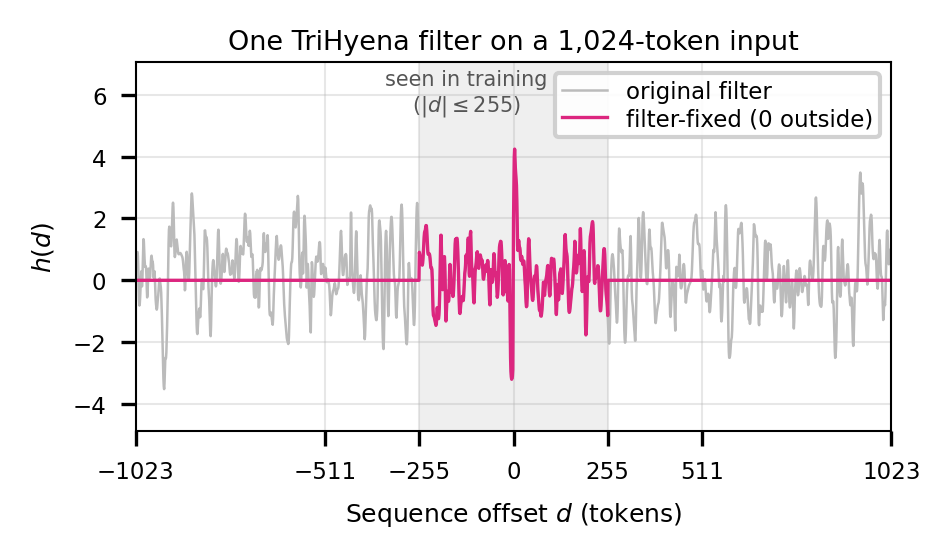}
\caption{One TriHyena filter of the \armFull model evaluated at $L = 1{,}024$. Grey: the
filter as generated by SIREN (legend: ``original filter''). Pink: the filter used at inference after zero masking (legend: ``filter-fixed''), with values beyond the trained
offset range (shaded, $|d| \leq L_\mathrm{train}-1$) set to zero.}
\label{fig:rf3-filter}
\end{figure}

\paragraph{Zero masking of untrained offsets.}
TriHyena generates its convolution filter with a SIREN network evaluated on a fixed coordinate
grid: the tap at offset $d = j-k$ is $h_d = \mathrm{SIREN}_\theta(s)$ with normalized coordinate
$s = d/(L_\mathrm{cache}-1)$, where $\theta$ denotes the SIREN parameters and $L_\mathrm{cache}$ the number of filter taps.
For the training crop of $L_\mathrm{train} = 256$ tokens, $L_\mathrm{cache} = 2L_\mathrm{train} - 1 = 511$ (held at this value at inference), so during training the filter is applied only at
$|d| \leq L_\mathrm{train}-1 = 255$, i.e.\ $|s| \leq 0.5$. On a longer sequence the same network is queried at offsets
it was never trained on ($|s| > 0.5$). Zero masking removes these contributions at inference: filter values
at $|d| > L_\mathrm{train}-1$ are set to zero (Figure~\ref{fig:rf3-filter}), with no retraining. TriHyena then
acts as a convolution over a window of 511 offsets through the 48 stacked blocks. On the 264 test entries of 768 tokens or more (387 protein chains; Table~\ref{tab:rf3-fixes}), zero masking recovers most of the loss of both \armHyb and \armFull: from
0.56 to 0.57 lDDT for \armHyb and from 0.13 to 0.54 for \armFull.

\begin{table}[htbp] 
\centering 
\caption{Protein best-of-1 (BO1) lDDT on test entries of 768 tokens or more under zero masking.} 
\label{tab:rf3-fixes} 
\small 
\begin{tabular}{lccc} 
\toprule Setting & $\geq 768$ & $768$--$1{,}023$ & $\geq 1{,}024$ \\ 
\midrule 
Baseline (\armBase) & 0.57 & 0.59 & 0.51 \\ 
\armHyb, unmasked & 0.56 & 0.58 & 0.48 \\ 
\armHyb, zero masking & 0.57 & 0.59 & 0.50 \\ 
\midrule 
\armFull, unmasked & 0.13 & 0.16 & 0.03 \\ 
\armFull, zero masking & 0.54 & 0.56 & 0.47 \\ 
\bottomrule 
\end{tabular} 
\end{table}



\section{Genie2}
\label{app:genie2-training-eval}

\subsection{Genie2 Training Details} 
\label{app:genie2-training}

We trained two Genie2 \citep{lin2024genie2} variants that differed only in the operator used in the pair stack. The baseline used the original five TriMul layers, whereas the TriHyena variant replaced all five TriMul layers with TriHyena layers.

Both models were trained in fp32 on 587,677 representative structures from AFDB \citep{varadi2022afdb}, restricted to structures containing at most 256 residues and with pLDDT $\geq 80$, following the original Genie2 training protocol \citep{lin2024genie2}.

\subsection{Genie2 Evaluation}
\label{app:genie2-evaluation}

For each evaluated checkpoint, we generated five unconditional backbones at every integer length from 50 to 256 residues, yielding 1,035 structures per checkpoint. For each generated backbone, ProteinMPNN proposed eight sequences, each of which was independently refolded with ESMFold \citep{lin2023esmfold}. Each ESMFold-refolded backbone was aligned to its corresponding Genie2-generated backbone using TM-align \citep{zhang2005tmalign}. We selected the candidate with the lowest C$\alpha$ RMSD between the generated and refolded backbones, termed the self-consistency RMSD ($\mathrm{scRMSD}$), for evaluation.

A generated backbone was considered designable if its best refold satisfied both $\mathrm{scRMSD} \leq 2$\,\AA{} and mean pLDDT $\geq 70$. Designability was defined as the fraction of generated backbones satisfying these criteria:

\begin{equation}
\mathrm{Designability}
=
\frac{N_{\mathrm{designable}}}
     {N_{\mathrm{generated}}},
\end{equation}
where $N_{\mathrm{designable}}$ is the number of designable backbones and $N_{\mathrm{generated}}=1{,}035$.

To quantify structural diversity, we clustered the designable backbones using TM-align with single-linkage clustering and a TM-score threshold of 0.6. Diversity was defined as the number of resulting designable clusters relative to all generated backbones:

\begin{equation}
\mathrm{Diversity}
=
\frac{N_{\mathrm{cluster}}}
     {N_{\mathrm{generated}}},
\end{equation}
where $N_{\mathrm{cluster}}$ is the number of clusters among designable backbones.

For each evaluated checkpoint, the F1 score was computed as the harmonic mean of designability and diversity:

\begin{equation}
\mathrm{F1}
=
\frac{
2 \times \mathrm{Designability} \times \mathrm{Diversity}
}{
\mathrm{Designability} + \mathrm{Diversity}
}.
\end{equation}

Reported F1 values are means of the per-checkpoint F1 scores rather than values computed from the mean designability and diversity. TriHyena was trained for up to 48 epochs because of its faster run-time.

\begin{figure}[t]
\centering
\includegraphics[width=\linewidth]{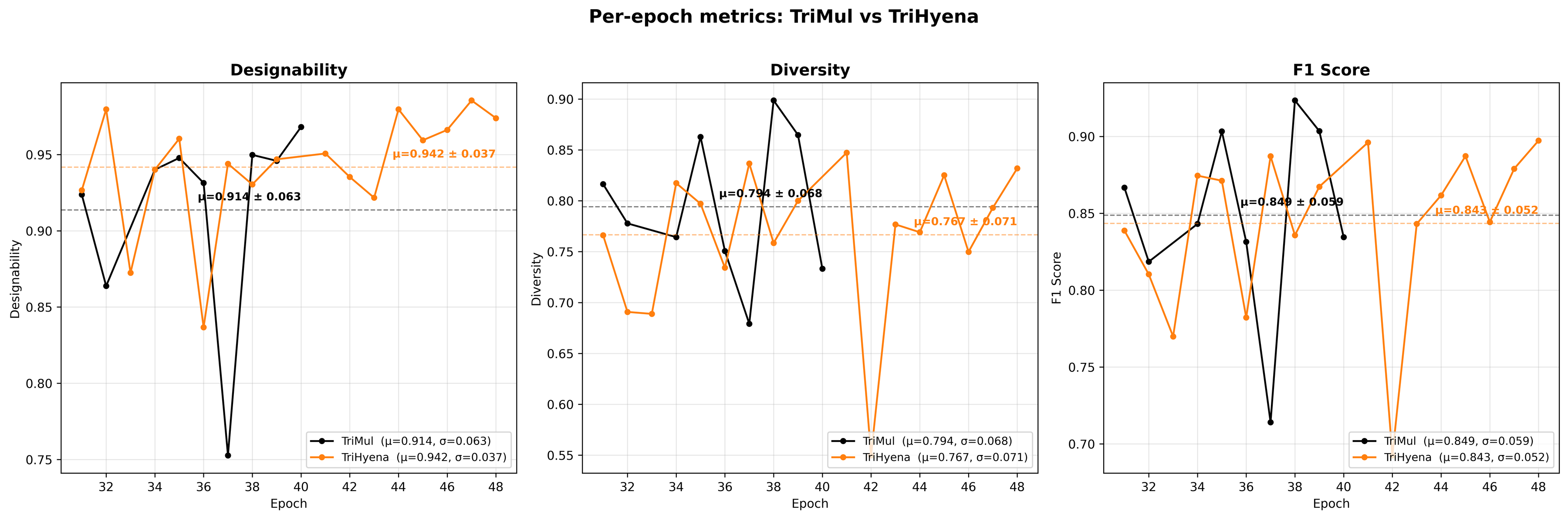}
\caption{Per-epoch designability, diversity, and F1 score for TriMul (black) and TriHyena (orange). Dashed lines indicate each method's mean across epochs, with mean $\pm$ standard deviation annotated above each line.}
\label{fig:trimul_trihyena_genie2}
\end{figure}

\end{document}